%% file: main.tex
\PassOptionsToPackage{dvipsnames}{xcolor}

\documentclass{article} 
\usepackage{iclr2027_conference,times}

\input{math_commands}

\input{preamble}

\usepackage{hyperref}
\usepackage{url}

\usepackage{xspace}
\newcommand{\method}{\textsc{DiscoDemo}\xspace}

\title{
DiscoDemo: \ourc{\uline{Disco}}vering Efficient and Diverse Robot \ourc{\uline{Demo}}nstrations for Imitation Learning
}

\author{
\begin{minipage}{0.95\textwidth}
\raggedright
\bfseries
\hypersetup{linkcolor=black}
Minho Park\textsuperscript{1,}\thanks{Equal Contribution} \quad
Kinam Kim\textsuperscript{1,\(*\)} \quad
Donghu Kim\textsuperscript{2} \quad
Byungkun Lee\textsuperscript{1} \quad
Dongyoon Hwang\textsuperscript{1} \\
Yongjae Shin\textsuperscript{1} \quad
Junha Hyung\textsuperscript{1} \quad
Hojoon Lee\textsuperscript{2,}\thanks{Corresponding Author} \quad
Jaegul Choo\textsuperscript{1,\(\dagger\)} \\
\vspace{2mm}
\normalfont
\textsuperscript{1}KAIST
\qquad
\textsuperscript{2}Holiday Robotics
\end{minipage}
}

\iclrfinalcopy 
\begin{document}

\maketitle
\lhead{}

\begin{abstract}
Behavior cloning for robotic manipulation is highly sensitive to the coverage of its training data, yet collecting diverse successful demonstrations remains costly.
We study reinforcement learning (RL) not as a deployable controller, but as a data generator for downstream visuomotor imitation learning.
Our key premise is that multiple successful solution modes cover a broader range of successful trajectories than repeatedly collecting a single narrow solution.
We introduce \method{}, which trains a privileged latent-conditioned RL generator with a parallel reverse curriculum and a success-gated diversity objective to discover diverse solutions while preserving task success.
The trained generator produces multiple successful closed-loop behaviors, which are safety-filtered and exported as a static dataset for downstream imitation learning.
Across four manipulation tasks, \method{} produces broader successful-state coverage and improves downstream visuomotor policies in both simulation and the real world.
Averaged across tasks, downstream success increases to $62\%$ from $40\%$ for the strongest baseline in simulation, and to $78\%$ from $39\%$ on the real robot, respectively.
These results suggest that successful solution diversity improves imitation learning for a pretrained flow-matching VLA and remains complementary to corrective data collection.\\
Project page: \url{https://davian-robotics.github.io/DiscoDemo/}
\end{abstract}

\vspace{-0.2cm}
\section{Introduction}
\label{sec:intro}
\vspace{-0.15cm}

Modern visuomotor policies are commonly trained from robot trajectories using behavior cloning (BC)~\cite{pomerleau1988alvinn}.
Scaling this paradigm, however, is constrained by the cost of physical data collection.
Robot trajectories are typically collected through human teleoperation, requiring access to physical robot hardware and operator time for every episode.
Consequently, even large robot datasets contain only on the order of a million episodes~\cite{o2024open,khazatsky2024droid}, far fewer than the billions of image--text pairs~\cite{schuhmann2022laion} or trillions of tokens~\cite{llama3} used to train foundation models in vision and language.

\vspace{-0.05cm}
Simulation offers a scalable alternative.
Robot trajectories can be generated cheaply and in parallel without requiring human teleoperation for every episode, and advances in sim-to-real transfer have made simulated experience increasingly useful for training real-world policies~\cite{torne2024reconciling,bauza2024demostart}.
Scaling simulated data, however, raises two key questions: \emph{how should successful trajectories be generated, and what distribution of successful behavior should they cover?}
Together, these choices determine the supervision available to downstream BC.

\vspace{-0.05cm}
Existing simulation pipelines generate successful trajectories in several ways.
Motion-planning methods construct feasible motions between task-relevant poses~\cite{chen2025robotwin,chen2025internvla,tian2026interndata}, but rely on task-specific planning structure.
MimicGen-style methods reduce this burden by retargeting successful demonstration segments to new scene configurations~\cite{mimicgen,skillmimicgen,jiang2025dexmimicgen}, but remain constrained by the behaviors contained in the source demonstrations and by where those behaviors can be successfully retargeted.
These limitations are particularly relevant in contact-rich manipulation, where geometric transformation alone may not preserve successful interactions as the scene changes.
Reinforcement learning (RL) provides a complementary approach by learning closed-loop behavior directly from task feedback, and recent methods bootstrap RL from demonstrations or pretrained behavior priors to make successful data generation practical under sparse rewards~\cite{rfcl,pld,xu2026expertgen}.

\begin{figure}[t]
\centering
\includegraphics[width=\linewidth]{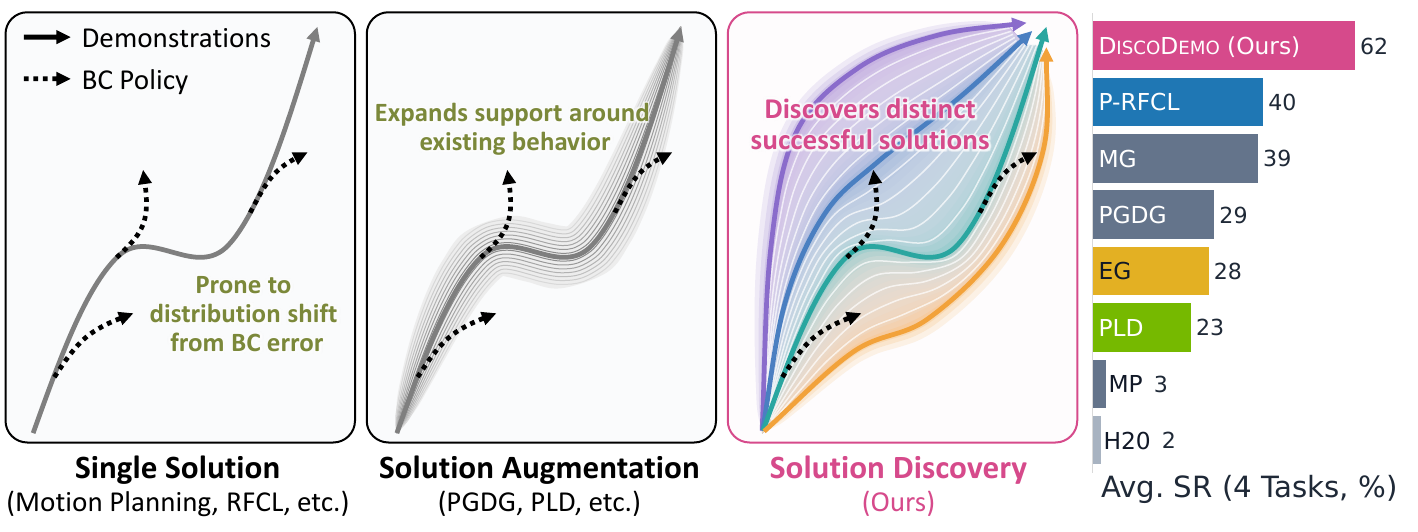}
\vspace{-4ex}
\caption{
\textbf{Solution discovery broadens coverage.}
Single-solution data provide narrow coverage, and augmentation mainly expands around existing trajectories.
\method{}, our method instead discovers distinct successful solutions that broaden state-space coverage.
}
\label{fig:motivation}
\end{figure}

\vspace{-0.05cm}
However, an RL policy that reliably solves the task does not necessarily generate a training distribution well suited to downstream BC.
As illustrated in \Cref{fig:motivation}, task reward alone can concentrate an RL policy around a narrow set of successful solutions, so collecting more rollouts need not substantially expand the states represented in the dataset~\cite{smerl}.
This limited coverage can make BC brittle when prediction errors move the learned policy beyond states supported by its training demonstrations~\cite{dagger}.
Perturbation- and recovery-based methods broaden supervision around an existing demonstration or policy~\cite{dai2026pgdg,pld}, but their additional experience remains anchored to the supplied behavior.
This motivates a different question: can the generator itself discover multiple successful ways of solving the task?

\vspace{-0.05cm}
We address this question with \method{}, which trains an RL generator to combine task competence with solution diversity.
Starting from a few successful demonstrations, we use a parallel reverse curriculum to bootstrap task-solving behavior under sparse rewards, using demonstration states for curriculum resets rather than treating demonstration actions as the target solution.
At the same time, we condition the generator on a latent skill and optimize a success-gated diversity objective that encourages different skills to realize distinct state-space behaviors while preserving task completion~\cite{smerl,park2024metra}.
The resulting generator discovers multiple successful solution modes whose union provides broader successful-state coverage for downstream BC.

\vspace{-0.05cm}
We evaluate \method{} on four real-world manipulation tasks reconstructed in simulation, spanning pick-and-place, stacking, and precision-sensitive peg insertion.
For each data source, we train the same $\pi_{0.5}$ vision-language-action policy~\cite{pi05}, pretrained on DROID~\cite{khazatsky2024droid}, and compare against human demonstrations and simulation data-generation methods based on planning~\cite{fu2026cap}, augmentation~\cite{mimicgen,dai2026pgdg}, and RL~\cite{pld,xu2026expertgen,rfcl}.
\method{} produces the strongest downstream policy across all four tasks in both simulation and the real world.
Averaged across tasks, it reaches $62\%$ success in simulation compared with $40\%$ for the strongest baseline, and $78\%$ compared with $39\%$ on the real robot.
The largest gains over the strongest baseline are $23$ percentage points on FMB-Round in simulation and $45$ on PnP-Banana in the real world.

\vspace{-0.3cm}
\section{Related Work}
\label{sec:rel-work}

\vspace{-0.3cm}
\paragraph{Simulation Data Generation for Behavior Cloning.}
A large body of work scales robot data by reusing or transforming successful behavior.
Motion-planning pipelines construct feasible trajectories from task structure~\citep{chen2025robotwin,chen2025internvla,tian2026interndata}, while MimicGen-style methods transfer demonstrations across new scene configurations~\citep{mimicgen,skillmimicgen,jiang2025dexmimicgen}.
DAgger~\citep{dagger}, DART~\citep{laskey2017dart}, and more recent intervention- and perturbation-based methods~\citep{hoque2024intervengen,dai2026pgdg} further expand coverage around learner errors and recovery states.
These approaches broaden scene, state, and recovery coverage from existing demonstrations or behaviors.
Our work follows a complementary line of RL-based data generation, training the generator itself to discover multiple successful modes.

\vspace{-0.3cm}
\paragraph{RL as a Data Generator.}
RL provides another route to simulation data generation by learning the action-generating policy from privileged state and task rewards.
RialTo~\citep{torne2024reconciling} uses simulation RL for visuomotor learning, whereas RLDG~\citep{xu2024rldg} uses real-world RL rollouts to fine-tune generalist policies.
RFCL~\citep{rfcl} and the contemporaneous DemoStart~\citep{bauza2024demostart} use closely related demonstration-driven curricula to bootstrap sparse-reward RL from successful states; DemoStart additionally distills the resulting privileged policy into a vision-based student.
PLD~\citep{pld} and ExpertGen~\citep{xu2026expertgen} connect RL data generation with pretrained downstream policies or behavior priors.
DLR~\citep{yang2025discover} instead discovers discrete behavioral patterns from successful demonstrations and reinforces them to construct diverse data for VLA pretraining.
Our work similarly studies the distribution produced by an RL generator, but discovers diverse successful solutions online with a latent-conditioned, success-gated objective.

\vspace{-0.3cm}
\paragraph{Diverse RL for Solution Discovery.}
Unsupervised skill discovery learns latent-conditioned policies with diverse behaviors without task-specific rewards~\citep{eysenbach2018diayn,sharma2019dynamics,laskin2022cic,park2022lipschitz,discodance,park2024metra}.
METRA~\citep{park2024metra} encourages latent skills to induce distinct state-space behaviors through a metric-aware representation, while task-aware methods such as SMERL~\citep{smerl} seek diversity without sacrificing task performance.
Our goal is not to learn a diverse repertoire for deployment, but to combine such diversity with trajectory-level success gating to broaden the range of successful trajectories available to downstream behavior cloning.

\begin{figure}[t]
\centering
\vspace{-0.1cm}
\includegraphics[width=\linewidth]{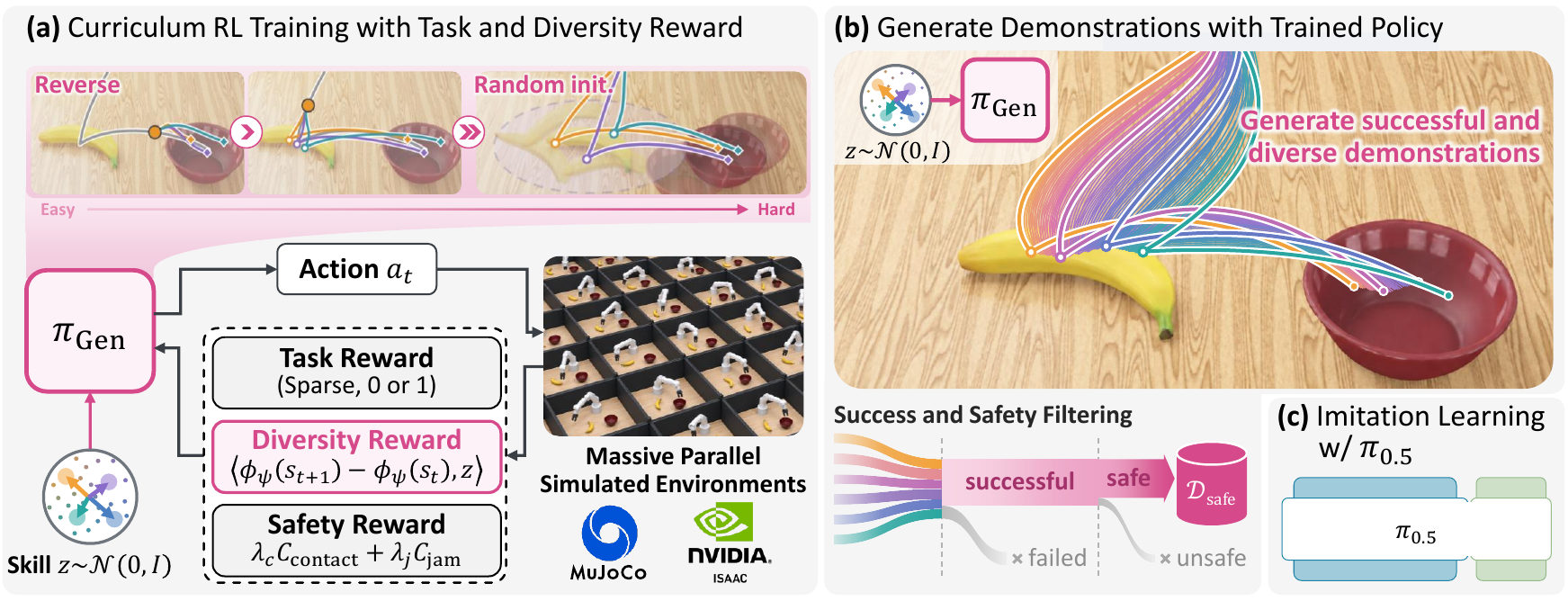}
\vspace{-0.6cm}
\caption{
\textbf{Overview of \method{}.}
(a) We train a privileged generator with a parallel reverse curriculum, sparse task reward, success-gated diversity, and safety shaping.
(b) The trained generator produces successful and diverse demonstrations, filtered for physically undesirable interactions.
(c) The retained observation--action data train a downstream visuomotor policy by imitation learning.
}
\label{fig:method}
\end{figure}

\vspace{-0.35cm}
\section{Method}
\label{sec:method}

\vspace{-0.2cm}
As illustrated in \Cref{fig:method}, we first train a latent-conditioned privileged policy as a demonstration generator, using curriculum RL with a task-aware diversity objective (\Cref{sec:method-train}).
We then freeze the generator and use it to produce diverse successful demonstrations, while filtering physically undesirable behaviors (\Cref{sec:method-generate}).
Finally, we use these filtered demonstrations to train a downstream visuomotor policy by behavior cloning,
without privileged information (\Cref{sec:method-il}).

\vspace{-0.3cm}
\subsection{Problem Setup}

\vspace{-0.15cm}
We assume access to a simulator with privileged state information and resets to a specified state $s\in\gS$.
The downstream visuomotor policy instead observes only $o=h(s)$, consisting of visual observations and robot proprioception.
For each task, we also assume a small set of human demonstrations
$\mathcal{D}_{\mathrm{human}}=\{\tau_i\}_{i=1}^{N}$, where
$\tau_i=\{(s_t,a_t)\}_{t=0}^{H_i-1}$.
We use these demonstrations as offline RL experience and their states as curriculum reset candidates.

\vspace{-0.05cm}
We model the underlying task as a finite-horizon Markov decision process (MDP),
$\mathcal{M}=\{\gS,\gA,\gT,r,\rho_0,\gamma\}$, with horizon $H$.
At each time step $t$, the data-generation policy
$\pi_{\mathrm{Gen}}(a_t | s_t;\theta_{\mathrm{Gen}})$
selects a continuous action $a_t\in\gA$, and the environment transitions to
$s_{t+1}\sim\gT(\cdot | s_t,a_t)$.
The initial state is sampled as $s_0\sim\rho_0$, and $\pi_{\mathrm{Gen}}$ is trained with a sparse task reward $r(s_t,a_t)$ to maximize
$\E_{\pi_{\mathrm{Gen}}}\![\sum_{t=0}^{H-1}\gamma^t r(s_t,a_t)]$.
We denote the downstream visuomotor policy by
$\pi_{\mathrm{IL}}(a_t | o_t;\theta_{\mathrm{IL}})$, which is trained only from the exported observation--action dataset.

\vspace{-0.2cm}
\subsection{Training a Task-Competent and Diverse Generator}
\label{sec:method-train}

\vspace{-0.05cm}
\paragraph{Efficient Learning with a Reverse Curriculum.}
With sparse task rewards, a policy starting from the original initial-state distribution may rarely encounter a success signal.
Following demonstration-based curricula~\citep{rfcl,bauza2024demostart}, we initially reset episodes to states near the end of successful demonstrations, so the policy only needs to complete the final part of the task.
As its success rate improves, we progressively move the reset boundary earlier in the demonstration, requiring the policy to solve increasingly longer portions of the task.

\vspace{-0.1cm}
We train $\pi_{\mathrm{Gen}}$ with FlashSAC~\citep{kim2026flashsac} using a massively parallel reverse curriculum.
For each demonstration $\tau_i$, we assign a group of parallel environments and maintain an independent curriculum boundary $c_i$, initialized near the end of the trajectory.
Episodes within each group are reset to states sampled from the demonstration suffix beginning at $c_i$.
When the success rate $\hat p_i$ over episodes started at demonstration $i$'s boundary reaches a threshold $p$, we move that boundary backward by $\Delta_c$:
\begin{equation}
c_i \leftarrow
\begin{cases}
\max(0,c_i-\Delta_c), & \hat p_i \ge p,\\
c_i, & \text{otherwise}.
\end{cases}
\end{equation}
Once half of the demonstrations are solved at $c_i=0$, all environments switch to the original task distribution $\rho_0$.
Thus, multiple demonstration-specific curricula advance concurrently from batched rollouts.
Details are provided in \Cref{sec:appn-rl-impl}.

\vspace{-0.2cm}
\paragraph{Discovering Multiple Successful Solutions.}
The curriculum helps the generator learn to solve the task, but task competence alone does not ensure that its demonstrations are useful for training a downstream visuomotor policy.
As illustrated in \Cref{fig:motivation}, under sparse rewards, once a reliable solution is discovered, the policy has little incentive to explore alternative solutions~\cite{smerl}, causing repeated rollouts to concentrate around a narrow state--action distribution.

\vspace{-0.05cm}
To encourage multiple successful solutions, we augment the task reward with a diversity reward based on METRA~\citep{park2024metra}.
We extend the generator to a latent-conditioned policy $\pi_{\mathrm{Gen}}(a_t | s_t,z;\theta)$, where a latent skill $z\sim\mathcal{N}(0,I)$ is sampled at the start of each episode and held fixed throughout the rollout.
Different values of $z$ encourage the generator to realize distinct successful behaviors across rollouts.
We jointly learn a state representation $\phi_\psi(s)\in\mathbb{R}^d$ from privileged simulator states, and reward state-space displacements that align with the sampled latent:
\begin{equation}
r_{\mathrm{div}}(s_t,s_{t+1},z)
=
\left\langle
\phi_\psi(s_{t+1})-\phi_\psi(s_t),z
\right\rangle.
\end{equation}

\vspace{-0.2cm}
Unconstrained diversity can also encourage distinct failure modes rather than useful task solutions.
Inspired by the structured-diversity principle of SMERL~\citep{smerl}, we therefore gate the diversity reward by trajectory success.
Let $y(\tau)\in\{0,1\}$ indicate whether trajectory $\tau$ completes the task.
We optimize the combined reward
\begin{equation}
\tilde r_t
=
r_{\mathrm{task}}(s_t,a_t)
+
\alpha\,y(\tau)\,
r_{\mathrm{div}}(s_t,s_{t+1},z),
\label{eq:combined-reward}
\end{equation}
where $\alpha$ controls the strength of the diversity objective.
Thus, the generator is rewarded for diversity only among trajectories that successfully complete the task.
Without the gate ($y(\tau)\equiv1$), the diversity reward dominates the sparse task reward and the generator fails to learn the task.

\vspace{-0.3cm}
\paragraph{Safety Shaping.}
Even successful trajectories can involve undesirable physical interactions, such as unintended scene contact or excessive force during insertion.
We therefore define a safety penalty
\begin{equation}
r_{\mathrm{safe}}(s_t,a_t)
=
\lambda_{\mathrm{c}}\,C_{\mathrm{contact}}(s_t)
+
\lambda_{\mathrm{j}}\,C_{\mathrm{jam}}(s_t),
\label{eq:safety-penalty}
\end{equation}
where $C_{\mathrm{contact}}$ penalizes undesirable robot--scene contact outside the intended finger--object interaction, and $C_{\mathrm{jam}}$, used for insertion tasks, penalizes excessive force between the held object and the insertion surface.
The generator is trained with the final reward $\tilde r_t-r_{\mathrm{safe}}(s_t,a_t)$, which softly biases the successful and diverse solutions toward physically plausible interactions.

\vspace{-0.15cm}
\subsection{Generating Successful and Diverse Demonstrations}
\label{sec:method-generate}

\vspace{-0.15cm}
\paragraph{Latent-Conditioned Rollouts.}

With the generator trained as described above, we freeze $\pi_{\mathrm{Gen}}$ and collect trajectories from initial states sampled from the original task distribution $\rho_0$.
For each rollout $i$, we independently sample a skill $z^{(i)}\sim\mathcal{N}(0,I)$ and hold it fixed until the episode ends.
The rollout yields a trajectory $\tau^{(i)}=\{(s_t^{(i)},a_t^{(i)})\}_{t=0}^{T_i-1}$ of length $T_i$.
By varying $z$ across rollouts, we collect different task-solving behaviors from the same trained generator.

\vspace{-0.3cm}
\paragraph{Safety Filtering and Dataset Export.}
Despite the safety penalty used during training \Cref{eq:safety-penalty}, generated trajectories may still contain undesirable physical interactions. We therefore discard failed trajectories and filter successful demonstrations using the same safety signals employed during training. Specifically, we retain
\begin{equation}
\mathcal{T}_{\mathrm{keep}}
=
\left\{
\tau
\;\middle|\;
y(\tau)=1,\;
\mathcal{F}_{\mathrm{task}}(\tau)=1
\right\},
\label{eq:filter}
\end{equation}
where $\mathcal{F}_{\mathrm{task}}$ checks contact magnitude and duration, plus insertion force for insertion tasks, and container displacement (\Cref{sec:appendix-safety}).
For each retained trajectory, we render the downstream observation $o_t=h(s_t)$ and pair it with the commanded joint-position target after action processing and the gripper command.
The resulting observation--action pairs form the static dataset $\mathcal{D}$ used for downstream imitation learning.

\vspace{-0.2cm}
\subsection{Downstream Imitation Learning}
\label{sec:method-il}

\vspace{-0.15cm}
Given the generated dataset $\mathcal{D}$, we train a downstream visuomotor policy $\pi_{\mathrm{IL}}$ using standard imitation learning.
Unlike the privileged generator $\pi_{\mathrm{Gen}}$, $\pi_{\mathrm{IL}}$ takes visual observations and robot proprioception as input, without access to the privileged simulator state or latent skill.
We optimize
\begin{equation}
\theta_{\mathrm{IL}}^\star
=
\argmin_{\theta_{\mathrm{IL}}}
\E_{(o,a)\sim\mathcal{D}}
\left[
\ell_{\mathrm{IL}}
\left(
\pi_{\mathrm{IL}}(\cdot|o;\theta_{\mathrm{IL}}),
a
\right)
\right],
\label{eq:downstream-il}
\end{equation}
where $\ell_{\mathrm{IL}}$ denotes the imitation-learning objective of the downstream policy.
$\pi_{\mathrm{IL}}$ is trained solely on the fixed dataset, without further interaction with the environment or the generator.

\vspace{-0.3cm}
\section{Experiments}
\label{sec:experiments}

\vspace{-0.15cm}
We evaluate whether generating successful, diverse demonstrations improves downstream visuomotor imitation learning.
We describe the experimental setup, then investigate the following questions:

\vspace{-0.15cm}
\begin{enumerate}[
    label=\textbf{RQ\arabic*:},
    leftmargin=*,
    topsep=2pt,
    itemsep=1pt,
    parsep=1pt,
    partopsep=0pt
]
\item Does our generated data improve downstream imitation in simulation and on a real robot?
\item How do the generated datasets differ in solution diversity and coverage, and how are these properties related to downstream performance?
\item How does the diversity objective trade off task competence, solution diversity, and downstream imitation performance?
\item How efficiently can diverse demonstrations be learned and generated, and are their benefits complementary to on-policy correction?
\end{enumerate}

\begin{figure}[t]
\centering
\includegraphics[width=\linewidth]{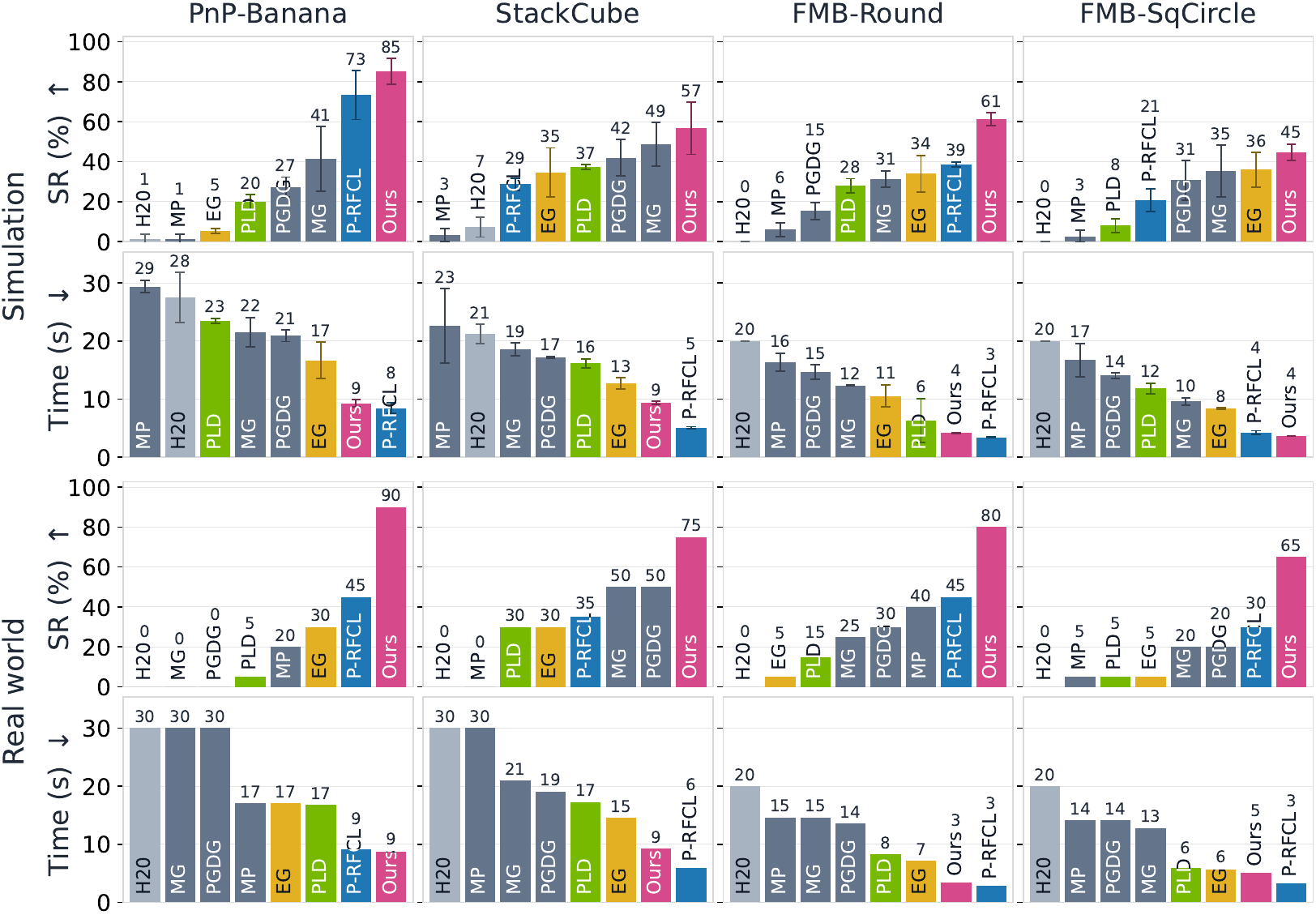}
\vspace{-0.7cm}
\caption{
\textbf{Downstream imitation performance in simulation and the real world.}
\method{} achieves the highest success rate across all tasks with competitive completion times.
Simulation: mean $\pm$ s.d.\ over three fine-tuning seeds.
}
\label{fig:main-results}
\end{figure}

\vspace{-0.2cm}
\subsection{Experimental Setup}
\label{sec:setup}

\vspace{-0.15cm}
\paragraph{Tasks.}
We evaluate on four Franka Research 3 manipulation tasks reconstructed in Isaac Sim~\citep{isaacsim}.
PnP-Banana places a banana into a bowl, StackCube stacks a red cube on a blue cube, and FMB-Round and FMB-SqCircle require inserting distinct peg geometries from the Functional Manipulation Benchmark (FMB)~\citep{fmb} into a fixture board.
All tasks randomize the initial object poses, with episode horizons of $30$\,s for PnP-Banana and StackCube and $20$\,s for the two insertion tasks (details in \Cref{sec:appn-tasks}).

\vspace{-0.3cm}
\paragraph{Data Sources and Baselines.}
We use the original $20$ human demonstrations as H20, while all generated-data methods provide $3{,}000$ successful demonstrations.
We compare H20 with motion planning (MP)~\citep{fu2026cap}, MimicGen (MG)~\citep{mimicgen}, PGDG~\citep{dai2026pgdg}, PLD~\citep{pld}, ExpertGen (EG)~\citep{xu2026expertgen}, P-RFCL, and \method{}.
For MP, we build a CaP-X-style scripted motion planner~\citep{fu2026cap}.
MG and PGDG transform and perturb human demonstrations, respectively, whereas PLD, EG, P-RFCL, and \method{} generate experience with RL.
P-RFCL (\method{}@$\alpha{=}0$) is our parallelized RFCL~\citep{rfcl} without diversity; PGDG reimplements its perturbation stage, while PLD and EG are reimplemented by state-based with FlashSAC (\Cref{sec:appn-baselines}).
\method{} uses a common diversity weight of $\alpha=0.3$ for the main downstream comparison.

\vspace{-0.3cm}
\paragraph{Downstream Training and Evaluation.}
For each generated-data method, we construct a dataset of $3{,}000$ successful demonstrations from left-shoulder and wrist cameras, and fine-tune the same DROID-pretrained $\pi_{0.5}$~\citep{pi05,khazatsky2024droid} visuomotor policy with joint-space action control using an identical training recipe.
We evaluate three seeds over $50$ episodes each in simulation and the preselected first seed over $20$ trials per task on the real robot, deployed zero-shot.
We report success rate (SR) and mean completion time over successful episodes (Time).
Further details are provided in \Cref{sec:appn-il}.

\vspace{-0.2cm}
\subsection{Downstream Imitation Performance}
\label{sec:main}

\vspace{-0.3cm}
\Cref{fig:main-results} compares downstream policies trained on each data source using the same visuomotor learner~\cite{khazatsky2024droid,pi05}.
Across both simulation and real-world evaluation, the results reveal distinct trade-offs in the types of demonstrations produced by different data-generation strategies.
H20 provides too little coverage from only a small set of human demonstrations, resulting in poor downstream success across the tasks.
MP~\cite{fu2026cap} substantially increases the amount of successful data, but remains relatively weak in our setting.
This challenge may be amplified by the joint-space action formulation of our DROID-pretrained $\pi_{0.5}$ downstream policy, which makes nominal planned trajectories more sensitive to imitation errors.

MG~\cite{mimicgen} and PGDG~\cite{dai2026pgdg} are considerably stronger in terms of success rate.
By transforming or perturbing human demonstrations, they preserve useful human behavior while expanding the range of training states.
Their successful rollouts, however, are often substantially longer, suggesting that these methods also inherit suboptimal motions from the source demonstrations.
For example, on PnP-Banana, the learned policy can take an unnecessarily circuitous path around the banana before grasping it.
We provide qualitative rollout comparisons illustrating these behaviors in \Cref{sec:appn-results}.

\vspace{-0.1cm}
P-RFCL exhibits the opposite behavior.
Because it uses the same RL generator as \method{} without the diversity objective, it often discovers a direct solution and, when successful, completes the task in the shortest time.
However, concentrating on such a narrow solution mode leaves the downstream policy with limited support once its rollout deviates from that nominal behavior, making recovery difficult and reducing overall success.
PLD~\cite{pld} and EG~\cite{xu2026expertgen} similarly demonstrate the benefit of RL-generated data for learning efficient motions, producing substantially shorter successful episodes than the human-augmentation baselines, but their downstream success rates remain limited.

In contrast, \method{} combines the two desirable properties.
Its diverse RL generator retains the short, goal-directed behaviors characteristic of RL-generated demonstrations while covering multiple successful solution modes that provide broader support for downstream execution and recovery.
Consequently, the same pattern holds in both simulation and on the real robot: \method{} achieves the strongest success rates while maintaining completion times close to the most efficient RL-based baselines.

\begin{figure}[t]
\centering
\includegraphics[width=\linewidth]{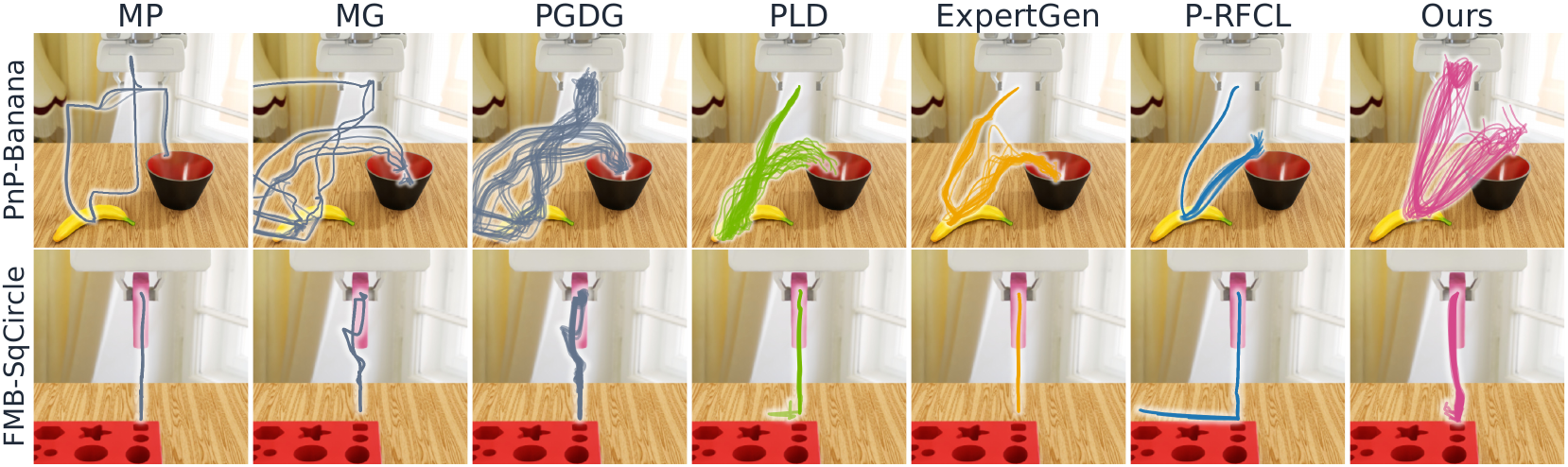}
\vspace{-0.6cm}
\caption{\textbf{Qualitative Results of Generated Demonstrations.} Each line traces the tool center point (TCP) of one successful demonstration from the same initial placement.}
\label{fig:qualitative}
\vspace{-0.4cm}
\end{figure}

\begin{figure}[t]
\centering
\includegraphics[width=\linewidth]{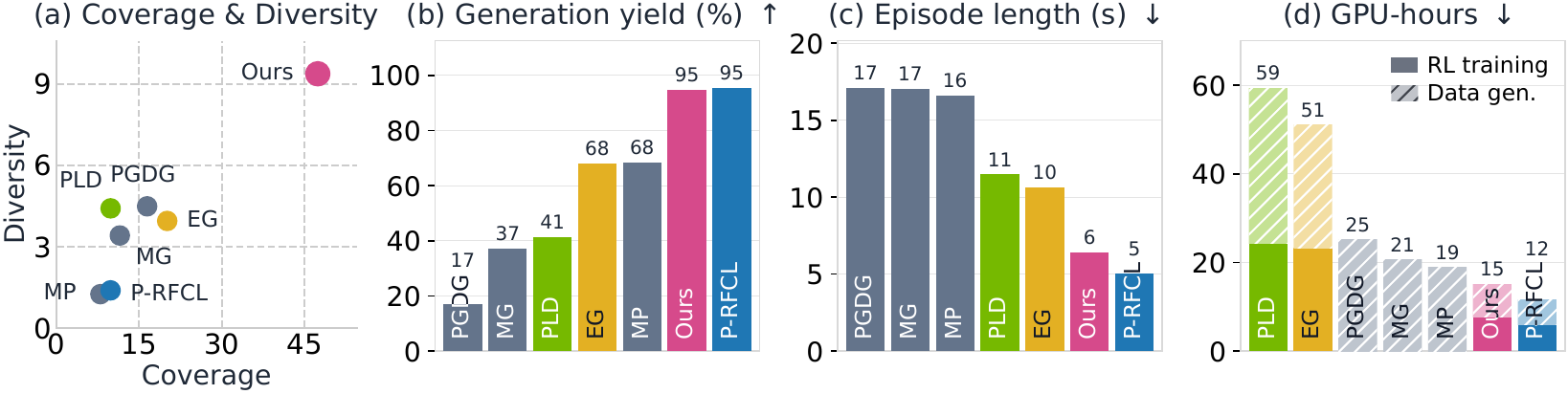}
\vspace{-0.6cm}
\caption{%
\textbf{Dataset analysis.}
On average across tasks, \method{} produces the most diverse and widest-coverage datasets, with high yield, short episodes, and low end-to-end GPU cost.
}
\label{fig:dataset-analysis}
\end{figure}

\vspace{-0.15cm}
\subsection{Analysis of Generated Demonstrations}
\label{sec:dataset-comparison}

\vspace{-0.1cm}
\paragraph{Qualitative Solution Diversity.}
To separate solution diversity from variation in initial conditions, we generate successful demonstrations from the same initial placement.
As shown in \Cref{fig:qualitative}, MP and P-RFCL repeatedly follow relatively narrow solution modes, whereas MG and PGDG mainly introduce variations around the source human behaviors.
PLD and ExpertGen exhibit greater variation through their pretrained behavior priors, but their solutions remain influenced by behaviors represented in those priors.
In contrast, \method{} produces distinct successful trajectories across all four tasks (\Cref{fig:qualitative}; StackCube and FMB-Round in \Cref{sec:appn-results}).
This suggests that the diversity objective expands the set of successful solution strategies, rather than only perturbing existing trajectories.

\vspace{-0.2cm}
\paragraph{Quantitative Diversity and Coverage.}
We quantify these differences in \Cref{fig:dataset-analysis}~(a) using the trajectory representation of PGDG~\cite{dai2026pgdg}.
Each successful trajectory is represented by its object and end-effector motion, and similarity between trajectories is computed with an RBF kernel.
We summarize each set using the Vendi score~\cite{friedman2023vendi}, which can be interpreted as the effective number of distinct trajectories (\Cref{app:div-cov}).
\emph{Diversity} applies this metric to successful rollouts from one fixed initial scene, isolating variation in how the same task instance is solved.
\emph{Coverage} instead applies it to the full $3{,}000$-demonstration dataset.

\vspace{-0.1cm}
\method{} achieves the highest diversity and coverage on average.
Dataset-level coverage is more closely associated with downstream success than fixed-scene diversity in our analysis (\Cref{sec:appn-metrics}).
Thus, visually distinct trajectories are not themselves the goal; multiple successful solutions are useful when they broaden the behaviors represented across the task distribution.

\vspace{-0.2cm}
\paragraph{Generation Yield and Trajectory Efficiency.}
The remaining panels characterize whether this broader support can be produced reliably and efficiently.
\emph{Generation yield} (\Cref{fig:dataset-analysis}~(b)) measures how often a data-generation attempt produces a successful trajectory before export filtering.
Yield alone does not determine dataset quality: scripted or transformation-based methods may simply reject many unsuccessful candidates, while for learned generators low yield can also indicate regions where the teacher has not acquired reliable task-solving behavior~\cite{mimicgen}.
Nevertheless, low yield restricts where successful supervision can be collected, even when the final dataset size is matched.
Despite optimizing for multiple solution modes, \method{} maintains high generation yield, indicating that its diversity does not simply arise from unconstrained exploration or additional failure modes.
\textit{Episode length} (\Cref{fig:dataset-analysis}~(c)) provides a complementary view of trajectory quality by measuring how directly successful rollouts reach the goal.
RL-based generators generally produce shorter, more goal-directed trajectories than methods that transform human demonstrations.
\method{} preserves this property while increasing solution coverage, whereas P-RFCL shows why short trajectories alone are insufficient: it often produces efficient successes around a much narrower solution distribution.

\vspace{-0.3cm}
\paragraph{End-to-End Data-Generation Cost.}
The GPU-hour panel (\Cref{fig:dataset-analysis}~(d)) measures the end-to-end compute required to obtain the final $3{,}000$ demonstrations, including generator training when applicable and subsequent rollout and rendering.
For a consistent comparison, all costs are normalized to a single process on one GPU.
The hatched data-generation portion, however, consists of independent rollout and rendering jobs and can be parallelized across multiple processes in practice.
Because P-RFCL and \method{} use lightweight state-based policies that can be trained and rolled out in massively parallel simulation, the additional computation used for solution discovery remains modest relative to methods that repeatedly invoke large vision-based behavior priors.

\vspace{-0.3cm}
\paragraph{Effect of Diversity Strength.}
We vary the diversity weight $\alpha$ to examine whether increasing behavioral diversity directly improves downstream imitation.
As $\alpha$ increases, the generator produces increasingly distinct solution modes, confirming that the objective controls solution diversity.
However, averaged over tasks, downstream success peaks at an intermediate value: when the diversity objective becomes too strong, further behavioral variation begins to come at the expense of reliable task completion, consistent with \citet{luo2026geometric}.
These results indicate that the desired property is not maximum diversity itself, but broad coverage by behaviors that remain successful.
Per task, the peak ranges from $\alpha=0.1$ to $\alpha=1$ (FMB-Round), as shown in \Cref{sec:alpha-full}.

\begin{figure}[t]
\centering
\includegraphics[width=\linewidth]{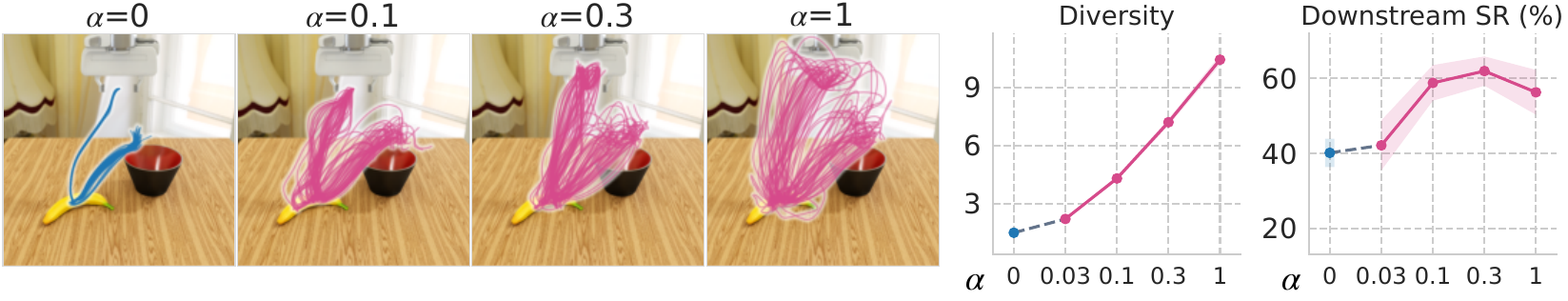}
\vspace{-0.7cm}
\caption{%
\textbf{Effect of diversity strength $\alpha$.}
Banana trajectory overlays and four-task averages diversity and SR. Diversity increases with $\alpha$, while downstream success peaks at an intermediate value.
}
\vspace{-0.2cm}
\label{fig:alpha-tradeoff}
\end{figure}

\begin{figure}[t]
\centering
\includegraphics[width=\linewidth]{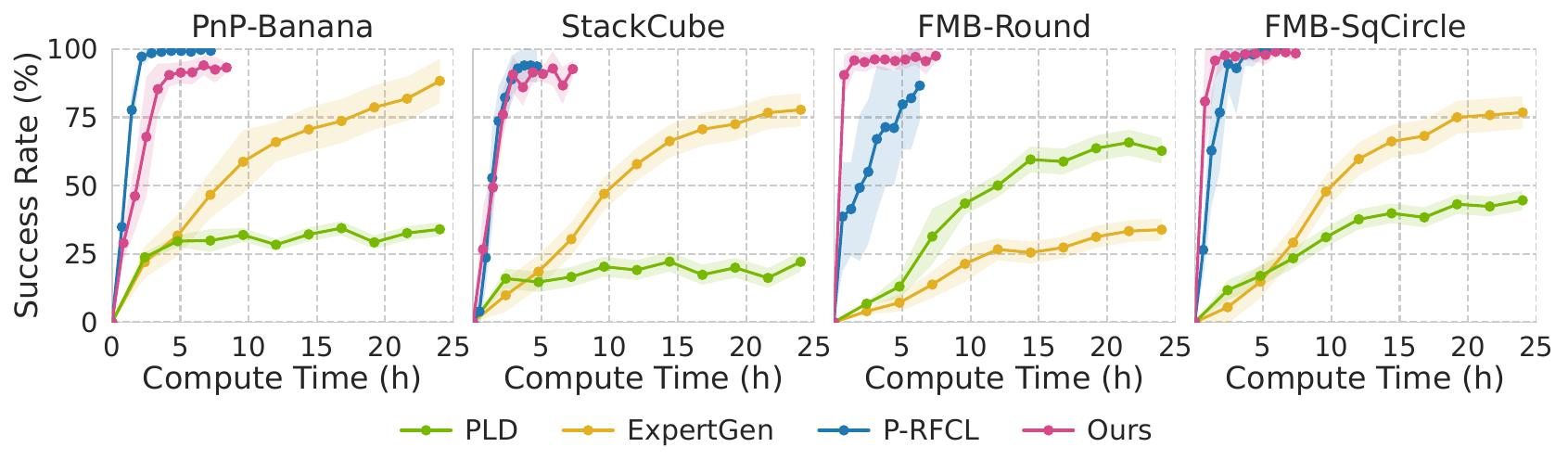}
\vspace{-0.6cm}
\caption{%
\textbf{Generator training efficiency.}
P-RFCL and \method{} complete training faster than PLD and ExpertGen. Times are estimated for the former and recorded for the latter (\Cref{sec:appn-efficiency}).
}
\label{fig:rl-baselines}
\vspace{0.1cm}
\end{figure}

\subsection{Generator Training Efficiency and Safety Curation}
\label{sec:exp-rl-training}

\vspace{-0.05cm}
\paragraph{Generator Training Efficiency.}
\Cref{fig:rl-baselines} compares training time, since per-step computation differs across generators.
PLD and ExpertGen rely on pretrained vision-based behavior priors and repeatedly invoke those models during RL, making each interaction expensive.
P-RFCL and \method{} instead train lightweight privileged-state policies with massively parallel simulation and therefore complete their training budgets faster.
The diversity objective adds representation learning and optimization, which lengthens training on every task, but the overhead remains small relative to the gap to vision-based generators.
In particular, on the insertion tasks, the broader exploration induced by the diversity objective can expose successful strategies early in training despite this additional computation.
Thus, the main efficiency advantage comes from the parallel state-based generator, while solution discovery adds only a task-dependent optimization overhead.

\vspace{-0.2cm}
\paragraph{Safety Shaping and Filtering.}
We examine how physically undesirable successful behaviors are removed before export using illegal contact (IC) and container displacement (CD) in \Cref{tab:safety-bs}.
IC measures unintended robot--scene contact, while CD measures motion of the bowl or fixture board during execution.
Safety shaping and filtering act at different stages: shaping softly biases the RL generator away from such interactions, whereas filtering rejects trajectories that violate the safety criteria after generation.
The ablation shows that shaping alone cannot eliminate all undesirable successes, since task completion can still outweigh a soft penalty.
Filtering produces the larger reduction in unsafe interactions, and combining both yields the lowest illegal contact on both tasks while reducing container displacement relative to the unconstrained generator; on FMB-SqCircle, the combination concedes some container displacement to filtering alone, as shaping concentrates on illegal contact.
By improving candidates before filtering, shaping also raises pass rates among successful rollouts from $31\%$ to $50\%$ on PnP-Banana and from $24\%$ to $46\%$ on FMB-SqCircle.

\vspace{-0.2cm}
\subsection{Complementarity with Corrective Data Collection}
\label{sec:exp-dagger}

\vspace{-0.1cm}
DAgger~\citep{dagger} and DART~\citep{laskey2017dart} also address behavior-cloning distribution shift by adding supervision beyond the demonstrations.
DAgger queries the teacher on learner-visited states, whereas DART perturbs teacher actions to expose recovery behavior (\Cref{sec:appn-baselines}).
In contrast, \method{} broadens the static dataset before downstream training by discovering multiple successful solution modes.
The approaches thus target different coverage: \method{} expands successful behaviors, while corrective collection adds supervision around off-nominal states.

\vspace{-0.1cm}
As shown in \Cref{fig:onpolicy}, both DAgger and DART substantially improve P-RFCL, confirming the value of corrective supervision.
Applying the same corrections to \method{} yields further gains, showing that solution diversity does not eliminate learner-specific distribution shift.
At the same time, \method{} remains stronger under the same correction, indicating that broader successful-behavior coverage remains useful after corrective data are added.
Unlike DAgger, DART requires no iterative learner rollouts.
Overall, \method{} improves the reusable static dataset without downstream interaction and remains complementary to corrective collection.

\begin{table}[t]
\centering
\begin{minipage}[t]{0.49\linewidth}
\centering
\caption{Effect of safety shaping and filtering \Cref{eq:filter}. Illegal contact (IC) and container displacement (CD) measured on the exported demonstrations.}
\label{tab:safety-bs}
\vspace{-0.15cm}
\setlength{\tabcolsep}{4pt}
\resizebox{\ifdim\width>\linewidth\linewidth\else\width\fi}{!}{
\begin{tabular}{@{}cc@{\hspace{8pt}}cccc@{}}
\toprule
\multicolumn{2}{c}{Generator} & \multicolumn{2}{c}{PnP-Banana} & \multicolumn{2}{c}{FMB-SqCircle} \\
\cmidrule(r{6pt}){1-2}\cmidrule(lr){3-4}\cmidrule(l){5-6}
Shaping & Filter & IC (N$\cdot$s) $\downarrow$ & CD (cm) $\downarrow$ & IC (N$\cdot$s) $\downarrow$ & CD (cm) $\downarrow$ \\
\drule
\xmark & \xmark & 21.3 & 1.37 & 73.0 & 10.48 \\
\cmark & \xmark & 15.1 & 1.32 & 22.9 & 13.13 \\
\xmark & \cmark & \underline{8.1} & \underline{1.06} & \underline{3.3} & \textbf{2.20} \\
\cmark & \cmark & \textbf{6.8} & \textbf{1.04} & \textbf{1.2} & \underline{5.89} \\
\bottomrule
\end{tabular}
}
\end{minipage}\hfill
\begin{minipage}[t]{0.49\linewidth}
\centering
\vspace{0.1cm}
\includegraphics[width=\linewidth]{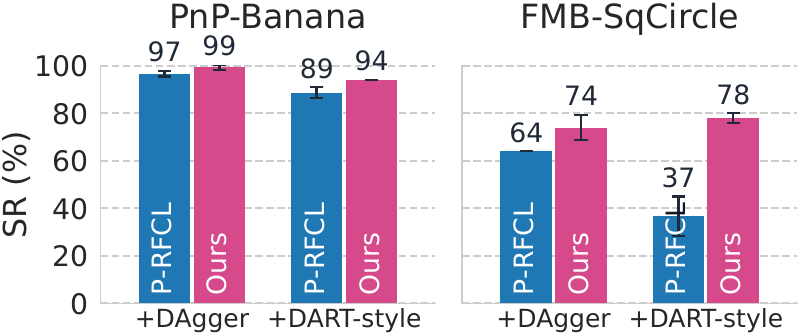}
\vspace{-0.6cm}
\captionof{figure}{Downstream success rate after corrective data collection.}
\label{fig:onpolicy}
\end{minipage}
\vspace{0.1cm}
\end{table}

\vspace{-0.3cm}
\section{Conclusion}
\label{sec:conclusion}

\vspace{-0.15cm}
We presented \method{}, which uses RL not as the controller, but as a generator of successful and diverse demonstrations for downstream visuomotor imitation learning.
By combining a parallel reverse curriculum with latent-conditioned solution discovery, \method{} expands the successful behaviors represented in the training data while retaining efficient state-based generation.
Across four tabletop tasks, these datasets improve the downstream performance of a pretrained flow-matching VLA in simulation and on the real robot.
Our analyses further show that broader dataset coverage is associated with stronger downstream performance, while excessive diversity can compromise task competence.
The gains remain complementary to corrective data collection such as DAgger and DART, with safety shaping and filtering providing an additional mechanism for curating generated trajectories.
Together, these results suggest that RL can be a practical tool for constructing broader imitation-learning datasets rather than only learning policies for direct deployment.

\vspace{-0.3cm}
\paragraph{Limitations.}
\method{} relies on accurate simulation, privileged state and reset access, and a well-defined success signal, limiting applicability to tasks with hard-to-model dynamics or ambiguous objectives.
Our evaluation is restricted to four tabletop tasks with simple scenes and a modest sim-to-real gap; more complex settings may require real-world adaptation.
The safety criteria are task-specific, and the diversity objective encourages distinct successful behaviors rather than directly optimizing downstream utility.
Future work could explore more general safety specifications and learner-aware objectives that target useful coverage more directly.

\subsection*{Reproducibility Statement}
We describe the simulated tasks, their initial-state distributions and success criteria, and the demonstrations in \Cref{sec:appn-tasks}; the generator's training procedure, reverse curriculum, diversity objective, safety terms, and all hyperparameters in \Cref{sec:appn-rl-impl}; our implementation of every baseline in \Cref{sec:appn-baselines}; the downstream fine-tuning recipe and evaluation protocol in \Cref{sec:appn-il}; and the definitions of all analysis metrics in \Cref{sec:appn-metrics}.
Simulated downstream results average three fine-tuning seeds with $50$ evaluation episodes each, generator training curves average five RL seeds, and real-robot results use $20$ trials per task and method.
The code, generated datasets, and trained checkpoints are available at \url{https://github.com/DAVIAN-Robotics/DiscoDemo}.

\subsection*{AI Use Statement}
In this work, we used generative AI tools to polish the writing, draft parts of the manuscript, and search for related work.
We also used them as coding assistants to implement the pipelines, baselines, and analysis and plotting scripts.
All AI-assisted text and code were reviewed by the authors, who take responsibility for the final content of this work.

\bibliography{iclr2027_conference}
\bibliographystyle{iclr2027_conference}

\clearpage

\appendix

\section*{Appendix}

In this appendix, we provide the implementation details needed to reproduce our results, together with additional results that do not fit in the main paper.
We first describe the tasks, their initial-state distributions and success conditions, and how the demonstrations were collected (\Cref{sec:appn-tasks}).
We then detail the generator, including its hyperparameters, the parallel reverse curriculum, the diversity objective with success gating, the safety terms, and its differences from the original RFCL (\Cref{sec:appn-rl-impl}), followed by our implementation of every baseline (\Cref{sec:appn-baselines}) and the downstream fine-tuning recipe and evaluation protocol (\Cref{sec:appn-il}).
Next, we define the metrics used to analyze the generated datasets, namely diversity, coverage, generation yield, and the safety metrics (\Cref{sec:appn-metrics}).
Finally, we report the full simulation and real-robot results, the diversity trade-off on every task, how training and generation cost are measured, and qualitative rollouts of the downstream policies (\Cref{sec:appn-results}).
Below is the table of contents for the appendix.

\startcontents[appendices]
\printcontents[appendices]{l}{1}{\setcounter{tocdepth}{2}}

\clearpage

\section{Tasks and Experimental Setup}
\label{sec:appn-tasks}

\begin{figure}[h]
\centering
\includegraphics[width=\linewidth]{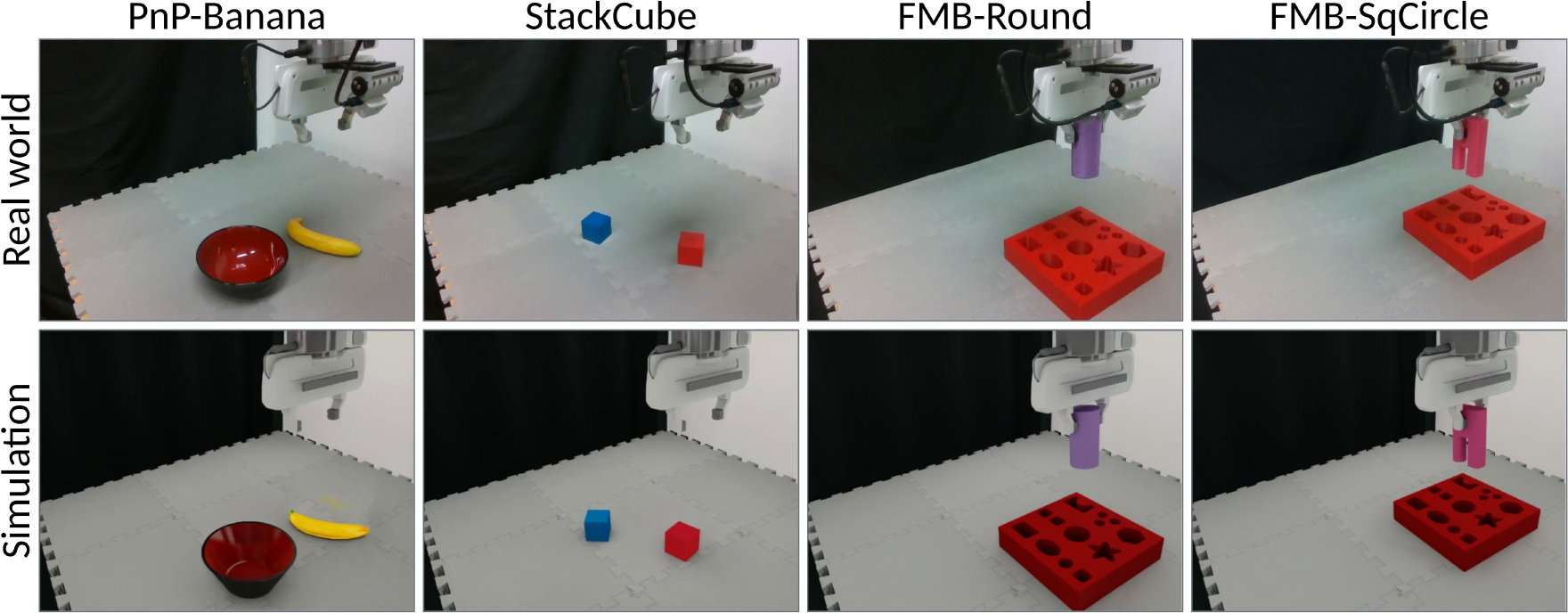}
\caption{\textbf{Experimental Setup.} The four tasks on the real FR3 (top) and in the simulated counterpart used for data generation (bottom), seen from the same exocentric camera.}
\label{fig:exp-setup}
\end{figure}

\paragraph{Robot and simulation.}
All tasks use a Franka Research 3 arm with a parallel-jaw gripper, reconstructed in Isaac Sim~\citep{isaacsim} from the real workcell (\Cref{fig:exp-setup}).
Physics runs at $120$\,Hz and the policy acts at $20$\,Hz.
The generator's action is $8$-dimensional in $[-1,1]$: seven relative joint-position increments, each scaled by a per-joint rate limit taken from the $95$th percentile of joint velocities in the teleoperated demonstrations and passed through a jerk-limited command filter, and one binary gripper command.
The privileged state is $45$-dimensional: the pose of the manipulated object and of the container (bowl, blue cube, or fixture board) together with their positions relative to the tool center point, the seven joint positions, the gripper state, the end-effector pose, the gripper actuation state, and the previous action.
The downstream policy instead observes a left-shoulder exocentric camera and a wrist camera, whose intrinsics, distortion, and extrinsics are calibrated on the real robot, together with the joint and gripper state.

\paragraph{Tasks.}
\Cref{tab:tasks} lists the horizon, initial-state distribution $\rho_0$, and success condition of each task.
An episode succeeds once the success condition in \Cref{tab:tasks} is met.
When training the generator, we additionally require the condition to hold for $20$ consecutive control steps ($1$\,s) before rewarding and terminating the episode; this training-time choice makes the generator settle into stable end states rather than pass through the goal, which yields more stable demonstrations.

\begin{table}[h]
\centering
\caption{Task specifications. Positions are in the robot base frame.}
\label{tab:tasks}
\small
\setlength{\tabcolsep}{4pt}
\begin{tabular}{@{}l c p{0.36\linewidth} p{0.33\linewidth}@{}}
\toprule
Task & Horizon & Initial-state distribution $\rho_0$ & Success condition \\
\midrule
PnP-Banana & $30$\,s & banana and bowl positions uniform in $x\in[0.41,0.64]$\,m, $y\in[-0.15,0.15]$\,m, at least $20$\,cm apart; banana yaw uniform in $\pm180^\circ$ & banana center inside the bowl, in contact with the bowl, and released by the gripper \\
StackCube & $30$\,s & red and blue cube positions in the same region, at least $15$\,cm apart; red-cube yaw uniform in $\pm45^\circ$ & red cube on the blue cube: height offset within $1$\,cm of $5$\,cm, horizontal offset $\le2.5$\,cm, speed $\le2$\,cm/s \\
FMB-Round & $20$\,s & peg grasped at the home configuration; fixture-board position uniform in $\pm3$\,cm and yaw in $\pm15^\circ$ & peg fully inserted (within $2$\,mm), lateral error $\le5$\,mm, axis misalignment $\le10^\circ$ \\
FMB-SqCircle & $20$\,s & as FMB-Round & as FMB-Round \\
\bottomrule
\end{tabular}
\end{table}

\paragraph{Demonstrations.}
For each task we collect $20$ demonstrations by teleoperating the real robot with a Meta Quest VR controller, recorded at $20$\,Hz.
The object poses of each episode are estimated from the exocentric RGB-D view, and the recorded joint trajectory is replayed in simulation from the corresponding initial scene; successful replays form the demonstration set used for the reverse curriculum and for H20, MimicGen, and PGDG.
For the insertion tasks, where the peg starts in the gripper, the gripper command is held closed.

\paragraph{Real-robot evaluation.}
Each policy is evaluated for $20$ trials per task on the real robot, reproducing the same set of initial object placements for every method.
Transfer is zero-shot: the policies are trained only on simulated data, without real-robot fine-tuning.
For every method we deploy the policy of the first training seed, fixed in advance of the real-robot evaluation.
SR and Time are defined as in simulation: SR is the fraction of the $20$ trials that succeed, and Time is the mean completion time over the successful trials.

\section{Generator Implementation}
\label{sec:appn-rl-impl}

\subsection{Training Setup}
We train P-RFCL and \method{} with FlashSAC~\citep{kim2026flashsac} for $50$M environment steps on a single B200 GPU, using $2{,}048$ parallel environments; PLD and ExpertGen are trained for up to $24$ hours each (\Cref{sec:appn-baselines}).
The actor and critic follow FlashSAC's residual MLP architecture with weight normalization; the critic is a pair of categorical (distributional) Q-functions whose target is the minimum of the two.
Privileged observations are mapped to $[-1,1]$ with fixed physical bounds---joint limits and the task workspace---rather than statistics of the demonstrations, so that no dimension saturates when the policy leaves the demonstrated region.
Rewards are scaled by a running estimate of the return scale.
Every batch draws half of its transitions from the demonstrations and half from online experience, throughout training.
\Cref{tab:hparams-rl} lists the hyperparameters.

\begin{table}[h]
\centering
\caption{Generator hyperparameters, shared by P-RFCL and \method{} and by all tasks.}
\label{tab:hparams-rl}
\small
\begin{tabular}{@{}ll@{}}
\toprule
Hyperparameter & Value \\
\midrule
Parallel environments & $2{,}048$ \\
Environment steps & $50$M \\
Actor / critic & $2$ residual blocks, width $128$ / $256$ \\
Critic support & $101$ atoms on $[-5,5]$, two critics \\
Optimizer & Adam, lr $3\!\times\!10^{-4}\to1.5\!\times\!10^{-4}$ (cosine) \\
Batch size & $2{,}048$ ($50\%$ demonstrations) \\
Gradient updates per vectorized step & $64$ \\
Actor update period & $2$ \\
$n$-step return, discount $\gamma$ & $8$, $0.99$ \\
Target update rate $\tau$ & $0.01$ \\
Initial temperature & $0.01$ \\
Target entropy & $\tfrac{|\mathcal{A}|}{2}\ln(2\pi e\,\sigma^2)$, $\sigma=0.15$ \\
Replay capacity & $100$M transitions \\
Precision & mixed (fp16), \texttt{torch.compile} \\
\midrule
Curriculum stride $\Delta c$ & $16$ demonstration steps \\
Advance threshold $p$ / window & $0.9$ / $10$ episodes \\
Reset-offset distribution & geometric, $q=0.5$ \\
Forward transition & half of the demonstrations solved \\
\midrule
Latent skill $z$ & $3$-dim, $\mathcal{N}(0,I)$ clipped to $[-1,1]$ \\
Representation $\phi_\psi$ & MLP $256$--$256$, ReLU, output dim.\ $3$ \\
$\phi_\psi$ / dual learning rate & $10^{-4}$ \\
Dual variable $\lambda$ (init.) / slack $\epsilon$ & $1.0$ / $10^{-3}$ \\
\midrule
Contact / jamming weight $\lambda_{\mathrm{c}}$ / $\lambda_{\mathrm{j}}$ & $10^{-4}$ / $10^{-6}$ \\
\bottomrule
\end{tabular}
\end{table}

\subsection{Parallel Reverse Curriculum}
\label{sec:appn-curriculum}
Let $\tau_1,\dots,\tau_N$ be the demonstrations ($N=20$) with lengths $T_i$, and let $c_i$ be the curriculum frontier of $\tau_i$, initialized near its end.
Whenever an environment resets during the reverse phase, it draws a demonstration and a start step jointly,
\begin{equation}
\Pr(i,t)\;\propto\;(1-q)^{\,t-c_i},\qquad t\in\{c_i,\dots,T_i-1\},\quad q=0.5,
\end{equation}
and resets the simulator to the demonstration state $s^{i}_{t}$, so that start steps concentrate at the frontiers and each demonstration is served by about $M/N\approx102$ parallel environments.
The curriculum of $\tau_i$ is advanced using the episodes of $\tau_i$ started exactly at its frontier: once at least $10$ such episodes have finished and the success rate $\hat p_i$ of the most recent $10$ reaches $p=0.9$, we set $c_i\leftarrow\max(0,c_i-\Delta c)$.
A demonstration is solved once this test also passes at $c_i=0$, and when half of the demonstrations are solved all environments switch to the task distribution $\rho_0$ (the forward phase).
Reverse-phase episodes are truncated after a fixed margin plus the remaining demonstration length $T_i-t$, and, as in the forward phase, count as successful only after the success condition has held for $1$\,s.

\subsection{Diversity Objective}
\label{sec:appn-diversity}
The latent skill $z\in\mathbb{R}^3$ is drawn from $\mathcal{N}(0,I)$ at every episode start and appended to the observations of the actor and the critics, clipped elementwise to $[-1,1]$.
The representation $\phi_\psi$ is an MLP with two hidden layers of $256$ ReLU units and output dimension $3$, applied to the normalized privileged state without the previous action and the gripper actuation state.
Following METRA~\citep{park2024metra}, $\phi_\psi$ maximizes $\E[(\phi_\psi(s')-\phi_\psi(s))^\top z]$ subject to $\tfrac{1}{d}\|\phi_\psi(s')-\phi_\psi(s)\|^2\le1$ with $d=3$ as in the official implementation, which we enforce with a dual variable $\lambda$ on the relaxed constraint $\min(\epsilon,\,1-\tfrac{1}{d}\|\phi_\psi(s')-\phi_\psi(s)\|^2)$; $\phi_\psi$ is trained on every sampled transition, including demonstrations.
Because the critic uses $n$-step returns, the diversity reward of a sampled transition is computed at update time over the same $n$ steps, $\langle\phi_\psi(s_{t+n})-\phi_\psi(s_t),z\rangle$, which is the sum of the per-step rewards in \Cref{eq:combined-reward}.

\paragraph{Success gating.}
The indicator $y(\tau)$ is known only when an episode ends, so it is applied retroactively.
Each episode receives a unique identifier that is stored with its transitions; when the episode terminates or is truncated, its outcome is written to a table indexed by that identifier, and each sampled transition looks up its gate at update time.
Transitions of episodes that have not finished yet, and demonstration transitions, receive no diversity reward.

\paragraph{Reward.}
The generator is trained on
\begin{equation}
r_t \;=\; r_{\mathrm{task}}(s_t,a_t) \;-\; r_{\mathrm{safe}}(s_t,a_t) \;+\; \alpha\,y(\tau)\,r_{\mathrm{div}}(s_t,s_{t+1},z),
\end{equation}
where $r_{\mathrm{task}}$ is $+1$ at the step at which the success condition has held for $1$\,s, which also terminates the episode, and $0$ otherwise.
The reward scaling above is applied to $r_{\mathrm{task}}-r_{\mathrm{safe}}$, and the diversity term is added after it without further scaling.
The diversity objective is active in both curriculum phases.
For the main downstream comparison, we use $\alpha=0.3$ on every task; P-RFCL uses the same generator with $\alpha=0$.

\subsection{Safety Shaping and Filtering}
\label{sec:appendix-safety}
The two terms of \Cref{eq:safety-penalty} are
\begin{align}
    C_{\mathrm{contact}}(s_t)
    &=
    \min\Bigl(\sum_{(b,o)\notin\mathcal{C}_{\mathrm{allow}}}\lVert F_{b,o}\rVert_2,\;c_{\mathrm{c}}\Bigr),
    \\
    C_{\mathrm{jam}}(s_t)
    &=
    \sum_{d\in\{\perp,\parallel\}}
    \bigl[\min(F^{d}_{\mathrm{peg}},\,c_{\mathrm{j}})-\delta_{d}\bigr]_+^2,
\end{align}
where $[x]_+=\max(x,0)$, so $C_{\mathrm{jam}}$ is a squared excess force (in N$^2$) that grows quadratically with how far the peg pushes past each deadband.
$C_{\mathrm{contact}}$ sums the contact forces $F_{b,o}$ between robot links $b$ and scene bodies $o$ over all pairs outside the allowed set $\mathcal{C}_{\mathrm{allow}}$, which contains only the finger--object contacts that constitute a grasp.
$C_{\mathrm{jam}}$ splits the force of the peg on the board into its axial component $F^{\perp}_{\mathrm{peg}}$ and lateral component $F^{\parallel}_{\mathrm{peg}}$ with separate deadbands $\delta_{\perp}<\delta_{\parallel}$: a correct insertion is guided by lateral contact with the hole wall, whereas a jammed peg pushes axially, so penalizing $\lVert F\rVert$ as a whole would favor the failure.
Both forces are saturated at $c_{\mathrm{c}}=c_{\mathrm{j}}=100$\,N so that rare force spikes do not dominate the reward.
We use $\lambda_{\mathrm{c}}=10^{-4}$ on every task and $\lambda_{\mathrm{j}}=10^{-6}$ with $\delta_{\perp}=5$\,N and $\delta_{\parallel}=15$\,N on the insertion tasks.
For the export filter we use $\epsilon_{\mathrm{j}}=(15\,\mathrm{N})^2$, i.e., we discard trajectories whose deadband-adjusted force vector has an $\ell_2$ norm above $15$\,N.

For contact, a trajectory is discarded if $C_{\mathrm{contact}}$ exceeds $\epsilon_{\mathrm{c}}=50$\,N at any step or remains above $20$\,N for $150$\,ms (three consecutive control steps).
The same export filter is applied to all datasets in the main comparison, with $C_{\mathrm{contact}}$ used for every task and $C_{\mathrm{jam}}$ additionally applied to the two insertion tasks.

\subsection{RFCL and Our Parallel Reimplementation}
\label{sec:appn-rfcl}

\paragraph{Original RFCL.}
RFCL~\citep{rfcl} targets sample- and demonstration-efficient RL: from a handful of demonstrations, it learns a single state-based policy that solves a sparse-reward task, and it is evaluated by the success rate reached per environment interaction.
Its reverse curriculum pins one environment to each demonstration (eight environments in total), samples start steps from a short geometric window at the demonstration's frontier, and moves the frontier back by four steps once the pinned environment has succeeded in its last three episodes.
A forward curriculum then samples initial states by prioritized level replay, and the policy is trained with SAC using two layer-normalized critics and a discount of $0.9$.

\paragraph{RFCL was not designed for data generation.}
RFCL optimizes the success of one policy and has no mechanism that keeps several successful behaviors, which is what \method{} adds with the diversity objective.
Its curriculum is also built for a few environments: the frontier of a demonstration advances only when every recent episode of its pinned environment succeeds, a criterion suited to measuring sample efficiency that does not scale to thousands of parallel environments.

\paragraph{Our reimplementation.}
We keep RFCL's idea of resetting to demonstration states but replace most of its components to exploit massively parallel simulation (\Cref{tab:rfcl-diff}); the change that makes the curriculum parallelizable is the advance criterion, which pools the outcomes of all episodes started at a demonstration's frontier instead of requiring each pinned environment to succeed (\Cref{sec:appn-curriculum}).
Because of these changes, the $\alpha=0$ results in \Cref{fig:alpha-tradeoff,fig:alpha-full} should not be compared directly with the numbers reported by \citet{rfcl}.

\begin{table}[h]
\centering
\caption{Our $\alpha=0$ generator (P-RFCL in our experiments) compared with the original RFCL configuration.}
\label{tab:rfcl-diff}
\small
\begin{tabular}{@{}lll@{}}
\toprule
& Original RFCL & Ours \\
\midrule
Reverse-phase environments & $8$ & $2{,}048$ \\
Environment-to-demonstration & one pinned environment each & a group of $\approx M/N$ environments each \\
Advance criterion & $3$ consecutive successes & success rate $\ge0.9$ over $10$ episodes \\
Curriculum stride & $4$ & $16$ \\
Forward phase & PLR over initial states & $\rho_0$ once half the demonstrations are solved \\
Success & at a single step & held for $1$\,s \\
Critic & MLP Q-function & categorical, residual \\
Discount / $n$-step & $0.9$ / $1$ & $0.99$ / $8$ \\
Batch size / updates per step & $256$ / $16$ & $2{,}048$ / $64$ \\
Target rate / initial temperature & $0.005$ / $1.0$ & $0.01$ / $0.01$ \\
Additional terms & --- & safety shaping \\
\bottomrule
\end{tabular}
\end{table}

\paragraph{Speedup from parallelization.}
We compare the two curricula on an earlier version of PnP-Banana with five demonstrations, running only the reverse phase on a single GPU with the same number of gradient updates per vectorized step.
With RFCL's rule of one pinned environment per demonstration, the reverse phase takes $8.2$ hours of wall-clock time, and giving the same rule $2{,}048$ environments does not help, since its per-environment criterion must then hold for hundreds of environments at once and the frontier stalls.
Our pooled criterion with $2{,}048$ environments completes the reverse phase in $29$ minutes, about $17\times$ faster in wall-clock time.
The gain is in wall-clock time rather than in samples: the parallel curriculum consumes more environment steps, which massively parallel simulation makes cheap.

\section{Baselines}
\label{sec:appn-baselines}
All generated-data baselines produce $3{,}000$ successful demonstrations under the same initial-state distribution, success condition, horizon, and camera configuration as \method{}.

\paragraph{H20.}
The $20$ teleoperated demonstrations of \Cref{sec:appn-tasks}, used directly.

\paragraph{Motion planning (MP, CaP-X-style).}
Following CaP-X~\citep{fu2026cap}, a scripted waypoint program reads the privileged object poses, solves inverse kinematics for each waypoint, and connects the resulting joint targets with jerk-limited trajectories under the same joint rate limits as the generator.
For PnP-Banana it grasps, lifts, and releases the banana above the bowl rim; for StackCube it adds a pre-grasp alignment waypoint and uses the cube's $90^\circ$ symmetry to choose the grasp; for the insertion tasks it aligns the grasped peg over the hole, descends, and inserts, retrying with a small spiral search if the peg jams.
The planner does not reason about collisions.

\paragraph{MimicGen (MG).}
Following MimicGen~\citep{mimicgen}, each demonstration is split at its first grasp into object-centric segments (grasp and place for PnP-Banana and StackCube, a single insertion segment for the insertion tasks).
For a new scene, each segment is transformed into the frame of its reference object, $T'_{\mathrm{ee}}=T'_{\mathrm{obj}}T_{\mathrm{obj}}^{-1}T_{\mathrm{ee}}$, and converted to joint targets by inverse kinematics; the source of each segment is chosen among the three demonstrations whose object layout is closest to the new scene.
Segments are joined by jerk-limited joint interpolation, and the gripper replays the recorded commands.

\paragraph{PGDG (perturbation-stage reimplementation).}
Since no official implementation of PGDG~\citep{dai2026pgdg} is available, we reimplement its perturbation stage on top of the MimicGen demonstrations.
For each source demonstration, the arm's joint-target sequence is perturbed by a residual B-spline with $16$ control points whose coefficients are Gaussian with initial scale $0.03$; we sample $16$ candidates per iteration for $2$ iterations, refit the perturbation distribution to the successful candidates that stay within a tube around the source (quantiles $0.1$ and $0.8$ of the object-feature deviation), and keep up to three successful perturbations per source.
Because several perturbed variants are kept per source, the retained dataset covers fewer distinct initial scenes than a MimicGen dataset of the same size.

\paragraph{PLD (FlashSAC variant).}
Following PLD~\citep{pld}, a $\pi_{0.5}$ policy fine-tuned on the demonstrations serves as a frozen base policy, and a residual policy outputs a correction to its arm command, $a=\mathrm{clip}(a_{\mathrm{base}}+0.4\,a_{\mathrm{res}})$.
The original PLD learns the residual from images, which did not learn within our budget; following object-centric residual RL~\citep{kim2026objectcentric}, we therefore give the residual policy the privileged state while the base policy remains vision-based, and train it with FlashSAC.
The residual is trained for up to $24$ hours ($2.4$--$2.6$M environment steps); demonstrations are collected from rollouts of the combined policy.

\paragraph{ExpertGen (EG, FlashSAC variant).}
Following ExpertGen~\citep{xu2026expertgen}, an RL policy steers the initial noise of the same frozen $\pi_{0.5}$ prior, making one $32$-dimensional decision per $15$-step action chunk with magnitude bounded by $3$.
We train it with FlashSAC in place of FastTD3 under the sparse task reward, for up to $24$ hours ($3.2$--$4.0$M environment steps), and collect demonstrations from the steered policy.

\paragraph{P-RFCL (\method{}@$\alpha{=}0$).}
The generator of \Cref{sec:appn-rl-impl} with $\alpha=0$, trained and rolled out identically to \method{}; it keeps RFCL's reset-to-demonstration curriculum but parallelizes it (\Cref{sec:appn-rfcl}).

\paragraph{DAgger.}
For \Cref{fig:onpolicy}, we use the frozen P-RFCL or \method{} generator as the teacher in DAgger~\citep{dagger}.
The learner starts from a policy fine-tuned on that generator's demonstrations and controls the robot during collection.
At each visited state, the teacher supplies an action label, converted to a joint-position target through its command filter.
We retain both successful and unsuccessful trajectories, add the labeled observations to the accumulated dataset, and fine-tune the learner before the next collection round.

\paragraph{DART.}
For DART~\citep{laskey2017dart}, the teacher controls the robot with perturbed actions.
We add fixed isotropic Gaussian noise with standard deviation $\sigma=0.3$ to the seven normalized arm commands, clip them to $[-1,1]$, and independently flip the gripper command with probability $0.02$.
Each observation is labeled with the noise-free teacher target, computed from the joint and command-filter state before executing the perturbed action.
The resulting demonstrations are filtered using the task-specific export criteria and used for downstream fine-tuning.

\section{Downstream Policy Training and Evaluation}
\label{sec:appn-il}

\paragraph{Dataset export.}
Each retained trajectory is rendered by resetting the simulator to its stored states and capturing the two cameras at $480\times640$ resolution and $20$\,Hz.
For \method{}, a new latent skill is drawn for every episode and held fixed throughout the rollout.
The observation stores the seven joint positions and the gripper state; the action stores the commanded joint-position target after action processing and a binary gripper command.

\paragraph{Policy training.}
We fine-tune the same DROID-pretrained $\pi_{0.5}$~\citep{pi05,khazatsky2024droid} on every dataset with the recipe in \Cref{tab:hparams-il}.
Images are resized with padding to $224\times224$ without augmentation, joint actions are expressed relative to the joint state at the start of each chunk while the gripper stays absolute, and states and actions are normalized by their $1$st and $99$th percentiles.
Each task uses one fixed language instruction, e.g., ``stack the red cube on the blue cube''.
For each dataset we train three policies with different seeds.

\begin{table}[h]
\centering
\caption{Downstream $\pi_{0.5}$ fine-tuning recipe, identical for every data source.}
\label{tab:hparams-il}
\small
\begin{tabular}{@{}ll@{}}
\toprule
Hyperparameter & Value \\
\midrule
Initialization & DROID-pretrained $\pi_{0.5}$ (joint-position actions) \\
Trained parameters & vision encoder, language model, action expert \\
Frozen parameters & action/time projections, multi-modal projector, LM head \\
Precision & bfloat16 \\
Optimizer & AdamW, $\beta=(0.9,0.95)$, weight decay $0.01$, grad.\ clip $1.0$ \\
Learning rate & $1$k warmup, cosine $2.5\!\times\!10^{-5}\to2.5\!\times\!10^{-6}$ \\
Batch size / steps & $16$ / $20$k \\
Cameras & left-shoulder exocentric, wrist; $224\times224$ \\
Action chunk & $15$ steps, all executed before replanning \\
Flow-matching steps & $10$ \\
\bottomrule
\end{tabular}
\end{table}

\paragraph{Simulation evaluation.}
Each policy is evaluated from its final checkpoint for $50$ episodes drawn from the task distribution $\rho_0$, the same distribution used to generate its training data.
An episode is counted as successful if the success condition is met at any step.
\emph{SR} is the fraction of successful episodes; \emph{Time} is the mean time to the first success over successful episodes, with a policy that never succeeds charged the full horizon $H$; and \emph{Succ./h} is the number of successes per hour of operation, $3600\cdot\mathrm{SR}/(\mathrm{SR}\cdot\mathrm{Time}+(1-\mathrm{SR})\cdot H)$, which charges every failure the full horizon.
Each metric is computed per training seed and reported as mean $\pm$ standard deviation over the three seeds.

\section{Analysis Metrics}
\label{sec:appn-metrics}

\subsection{Diversity and Coverage Metrics}
\label{app:div-cov}
We follow the trajectory representation of PGDG~\cite{dai2026pgdg} with two changes:
we drop the action channel and replace the log-determinant with the Vendi score.

\paragraph{Features.}
At each step we use $\psi(s_t)=[\mathbf{p}_{\mathrm{obj}},\mathbf{q}_{\mathrm{obj}},
\mathbf{p}_{\mathrm{ee}},\mathbf{q}_{\mathrm{ee}}]\in\mathbb{R}^{14}$, the single-arm version of
PGDG's $[T_{\mathrm{obj}},T_{\mathrm{ee}}]$; quaternions are sign-canonicalized ($w\ge0$).
PGDG additionally appends the action $a_t$. We omit it because action signals differ in character
across generators---replayed or planned trajectories are nearly constant, whereas RL policies are
not---so the action channel would dominate the comparison without reflecting different solutions.
When the end-effector pose is not stored, we recover it by forward kinematics
(max.\ position error $2.2\times10^{-5}$\,m against logged poses).

\paragraph{Embedding and kernel.}
Features are z-scored with statistics pooled over all sources of a task. Each trajectory is truncated
at its first success (for coverage, at the end of the $1$\,s success hold, since the exported datasets flag success only after the hold), linearly resampled to $64$ steps, and transformed per dimension with DCT-II;
we drop the DC term and keep the next $8$ coefficients (PGDG Eq.~31), giving a $112$-dimensional
embedding. Similarity is $k(\xi_i,\xi_j)=\exp(-\gamma\|\xi_i-\xi_j\|^2)$ (PGDG Eq.~32), with
$\gamma$ set per task to the inverse median pairwise squared distance over all sources,
as PGDG does not report its bandwidth.

\paragraph{Vendi score.}
For a set of $m$ trajectories with kernel matrix $K$, $\mathrm{VS}=\exp(-\sum_i\lambda_i\log\lambda_i)$,
where $\lambda_i$ are the eigenvalues of $K/m$; it ranges from $1$ (identical trajectories) to $m$.
PGDG's objective $\log\det(K+\epsilon I)$ is suited to subset selection but not to evaluation:
near-duplicate trajectories drive eigenvalues to zero, so the value is set by $\log\epsilon$ and
the number of such eigenvalues rather than by diversity. Because VS grows with $m$, we fix $m$ per
metric and average over random subsets.

\begin{figure}[t]
\centering
\includegraphics[width=\linewidth]{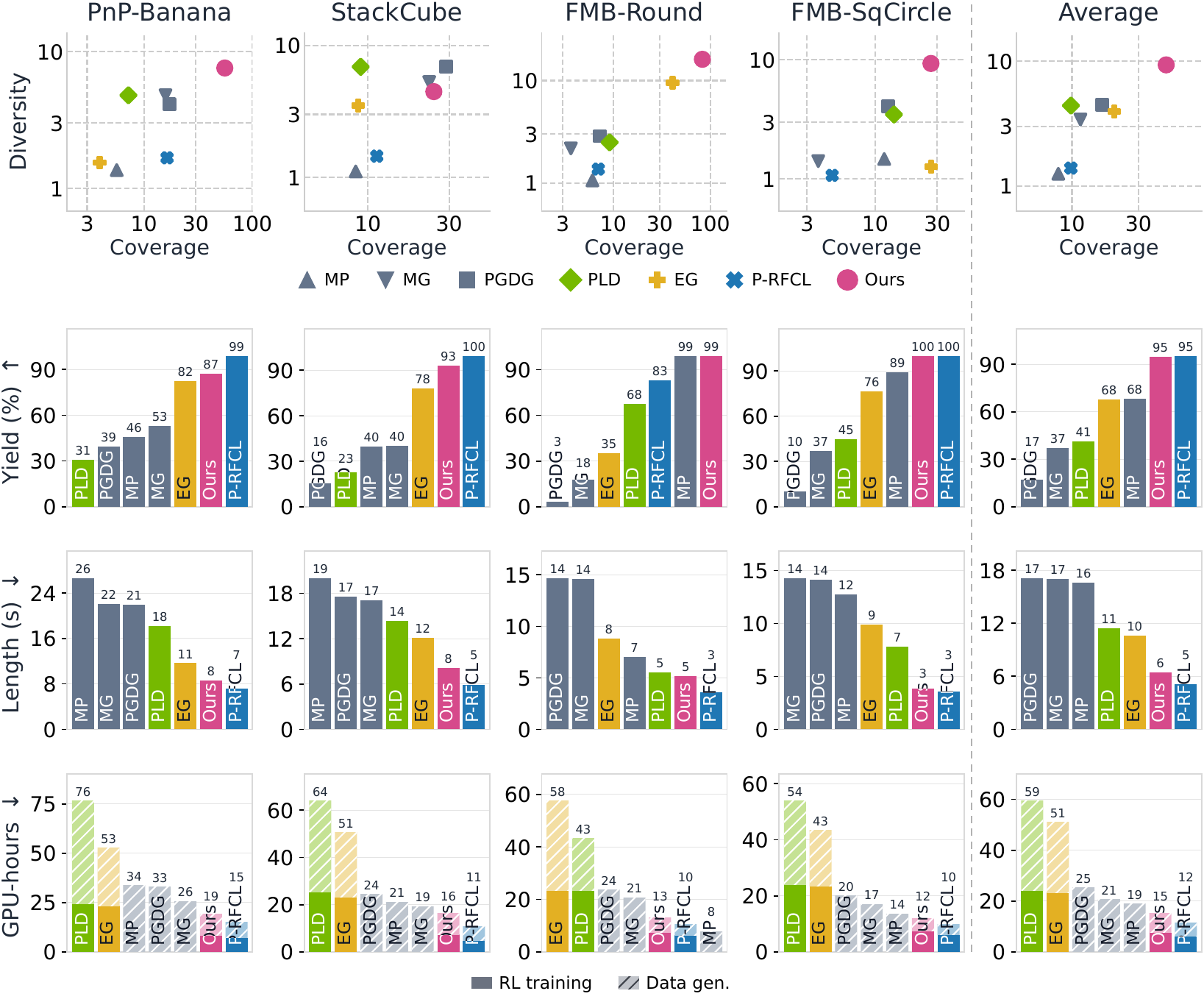}
\vspace{-0.6cm}
\caption{%
\textbf{Per-task dataset analysis.}
\method{} attains the highest diversity and the highest or tied-highest coverage on three of the four tasks; on StackCube, PGDG exceeds it on both.
}
\label{fig:dataset-taskwise}
\end{figure}

\paragraph{Inputs.}
\emph{Diversity}: for each task, one fixed initial scene; per source, $24$ successful rollouts,
$m{=}20$, averaged over $200$ subsets.
\emph{Coverage}: per source, $1{,}000$ trajectories sampled from the $3{,}000$-demonstration
training dataset (PLD and ExpertGen: from their generated pools), $m{=}200$, averaged over $100$ subsets.
Both are averaged over the four tasks.
\Cref{fig:dataset-taskwise} reports every panel of \Cref{fig:dataset-analysis} per task; its coverage--diversity row uses logarithmic axes because the baselines cluster within a narrow range on individual tasks.

The diversity axis of \Cref{fig:alpha-tradeoff,fig:alpha-full} applies the same diversity procedure to the rollouts of each generator in the $\alpha$ sweep, from the same fixed scene per task.

\begin{figure}[t]
\centering
\includegraphics[width=\linewidth]{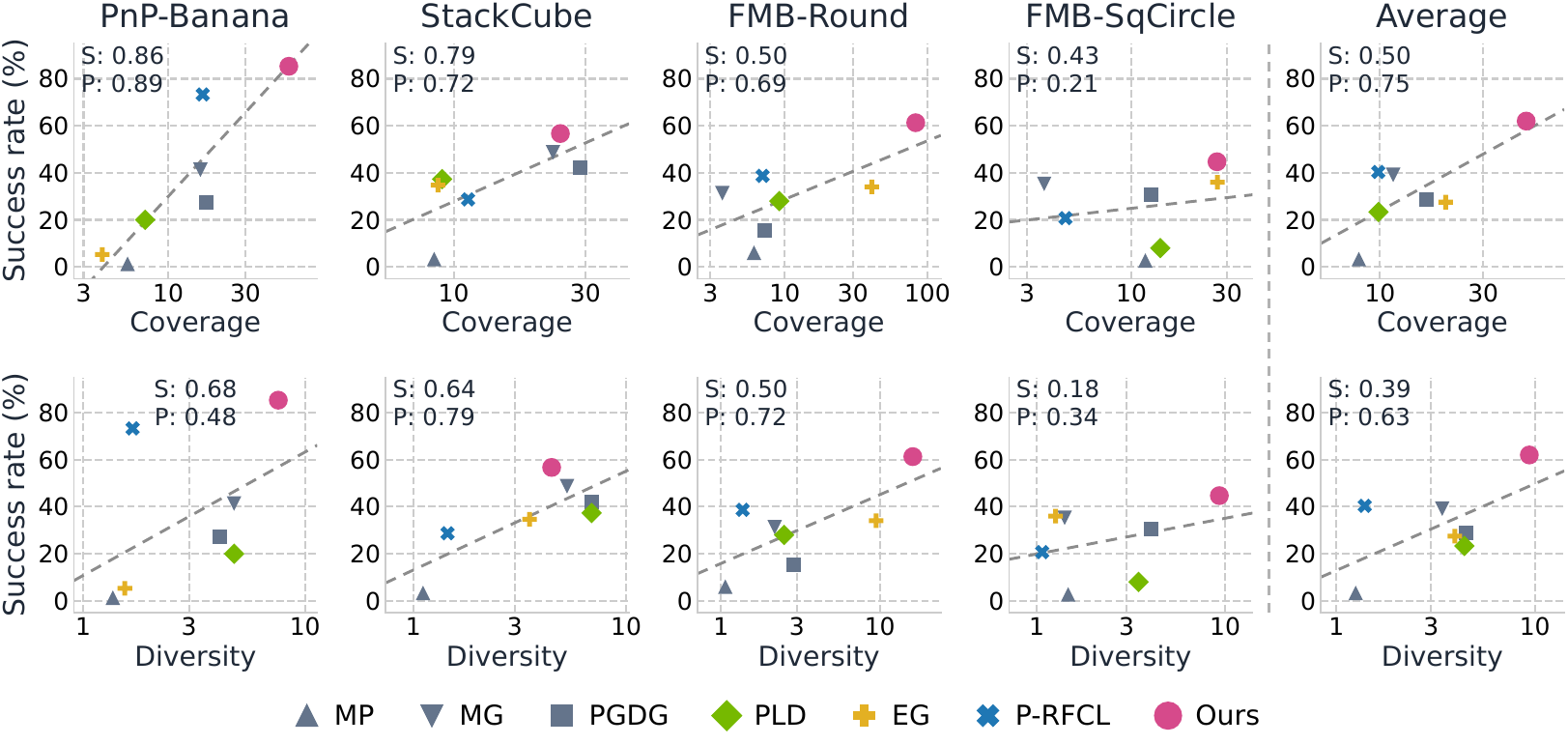}
\vspace{-0.6cm}
\caption{%
\textbf{Coverage and diversity versus downstream success.}
Dataset coverage (top) tracks the success of the downstream policy more closely than fixed-scene diversity (bottom).
}
\label{fig:cov-sr}
\end{figure}

\paragraph{Coverage, diversity, and downstream success.}
\Cref{fig:cov-sr} plots, for each generator, the coverage (top) and diversity (bottom) of its dataset against the success rate of the $\pi_{0.5}$ policy trained on it (\Cref{fig:main-results}).
Each panel reports the Spearman (S) and Pearson (P) correlation over the seven generators; Pearson coefficients and the dashed least-squares fits use the $\log_{10}$ of the metric.
Coverage of the full training dataset tracks downstream success more closely than diversity from a single fixed initial state: its rank correlation is higher or equal on every task (S $0.86$ vs.\ $0.68$ on PnP-Banana, $0.79$ vs.\ $0.64$ on StackCube, $0.50$ vs.\ $0.50$ on FMB-Round, $0.43$ vs.\ $0.18$ on FMB-SqCircle, and $0.50$ vs.\ $0.39$ on the task average).
These associations support dataset-level coverage as a useful indicator of downstream performance, rather than fixed-scene diversity alone; with seven generators per task, however, individual coefficients remain noisy.

\subsection{Generation Yield and Qualitative Trajectories}
\begin{figure}[t]
\centering
\includegraphics[width=0.95\linewidth]{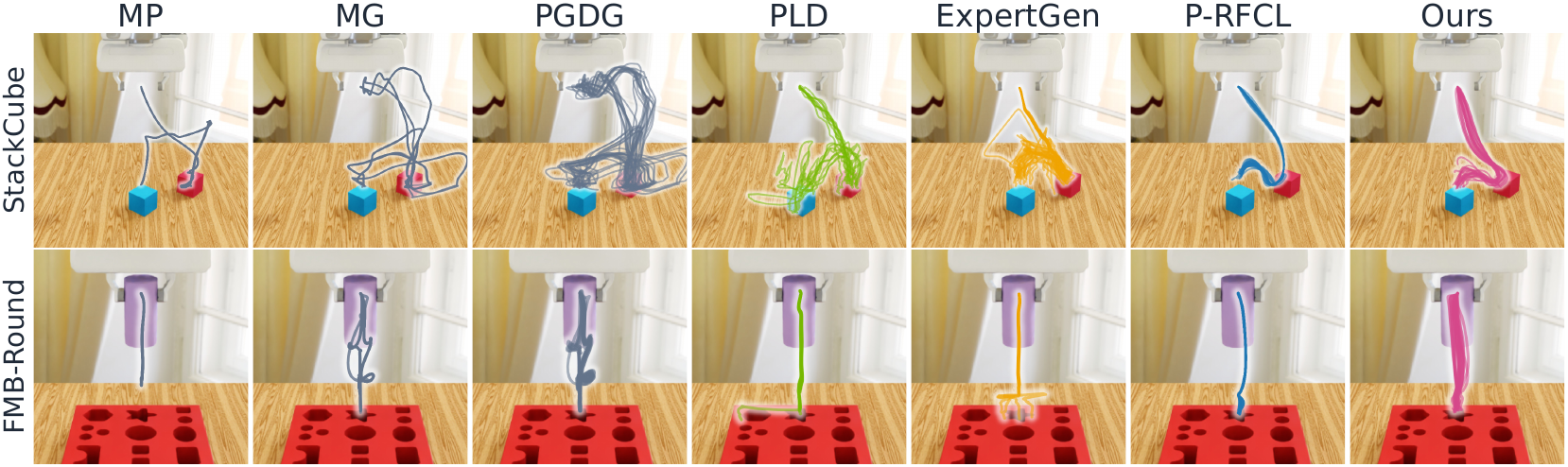}
\caption{Generated demonstrations on StackCube and FMB-Round, each line one successful demonstration from the same initial placement.}
\label{fig:qualitative-supp}
\end{figure}

\emph{Generation yield} is the fraction of rollouts from $\rho_0$ that succeed, before any export filter.
For PGDG it is the product of the MimicGen yield and the success rate of its perturbed candidates; for PLD and ExpertGen it is the success rate at the end of RL training.
The trajectory overlays in \Cref{fig:qualitative,fig:qualitative-supp,fig:alpha-tradeoff} use the same fixed scene per task as the diversity metric: for each method we collect $24$ successful rollouts and project the tool center point into the calibrated exocentric camera.

\subsection{Safety Metrics}
\label{sec:appn-safety-metrics}
We report illegal contact (IC) and container displacement (CD) on the exported demonstrations.
Let $\mathcal{C}_{\mathrm{allow}}$ denote the intended finger--object contacts.
For a trajectory $\tau$, IC is defined as

\begin{equation}
\mathrm{IC}(\tau)
=
\sum_t
\sum_{(b,o)\notin\mathcal{C}_{\mathrm{allow}}}
\left\|F_{b,o}(t)\right\|_2
\Delta t,
\qquad
\Delta t=0.05~\mathrm{s},
\end{equation}

where $F_{b,o}(t)$ is the simulator contact force between robot body $b$ and scene object $o$.
The allowed contacts are finger--banana contact for PnP-Banana and finger--peg contact for FMB-SqCircle; all other monitored robot--scene pairs are treated as illegal.
Unlike the training contact penalty, IC does not clip large contact forces and is used only as an evaluation metric.

Container displacement is defined as

\begin{equation}
\mathrm{CD}(\tau)
=
\max_t
\left\|
\mathbf{p}_{\mathrm{container}}^{xy}(t)
-
\mathbf{p}_{\mathrm{container}}^{xy}(0)
\right\|_2,
\end{equation}

where the container is the bowl for PnP-Banana and the fixture board for FMB-SqCircle.
We report the dataset mean of both metrics, with CD converted to centimeters.

\paragraph{Filter pass rates.}
\Cref{tab:safety-yield} lists how many successful rollouts of each generator in \Cref{tab:safety-bs} pass the export filter of \Cref{eq:filter}.
Each pool is collected with deterministic actions from a single checkpoint at a comparable training budget ($29$--$33$M environment steps).
Unfiltered rows of \Cref{tab:safety-bs} use a fixed random subset of $3{,}000$ successes, and filtered rows use the first $3{,}000$ passing successes in the same order.
Without shaping, only $2{,}701$ FMB-SqCircle successes pass, so that row of \Cref{tab:safety-bs} is computed on these $2{,}701$ trajectories.

\begin{table}[h]
\centering
\caption{Export-filter pass rates among successful rollouts. The last column is the number of successes needed for $3{,}000$ filtered demonstrations.}
\label{tab:safety-yield}
\resizebox{\ifdim\width>\linewidth\linewidth\else\width\fi}{!}{
\begin{tabular}{@{}llrrrr@{}}
\toprule
Task & Shaping & Successes & Passed & Pass rate (\%) & Needed \\
\midrule
PnP-Banana & \xmark & 11{,}428 & 3{,}577 & 31.3 & 9{,}585 \\
PnP-Banana & \cmark & 10{,}730 & 5{,}358 & 49.9 & 6{,}008 \\
\midrule
FMB-SqCircle & \xmark & 11{,}371 & 2{,}701 & 23.8 & 12{,}630 \\
FMB-SqCircle & \cmark & 11{,}632 & 5{,}395 & 46.4 & 6{,}468 \\
\bottomrule
\end{tabular}
}
\end{table}

\section{Additional Results}
\label{sec:appn-results}

\subsection{Main results --- full tables}
\label{sec:main-sim}

For completeness, \Cref{tab:main,tab:main-sim} report the real-robot and simulation results shown in \Cref{fig:main-results}.

\begin{table}[t]
\centering
\caption{Real-world experiments.}
\label{tab:main}
\small
\setlength{\tabcolsep}{2.5pt}
\resizebox{\textwidth}{!}{%
\begin{tabular}{@{}lcccccccccccc@{}}
\toprule
& \multicolumn{3}{c}{PnP-Banana} & \multicolumn{3}{c}{StackCube} & \multicolumn{3}{c}{FMB-Round} & \multicolumn{3}{c}{FMB-SqCircle} \\
\cmidrule(lr){2-4}\cmidrule(lr){5-7}\cmidrule(lr){8-10}\cmidrule(lr){11-13}
\textbf{Data source} & SR (\%) $\uparrow$ & Time (s) $\downarrow$ & Succ./h $\uparrow$ & SR (\%) $\uparrow$ & Time (s) $\downarrow$ & Succ./h $\uparrow$ & SR (\%) $\uparrow$ & Time (s) $\downarrow$ & Succ./h $\uparrow$ & SR (\%) $\uparrow$ & Time (s) $\downarrow$ & Succ./h $\uparrow$ \\
\drule
H20                                  & 0.0 & \textcolor{gray}{30.0} & 0 & 0.0 & \textcolor{gray}{30.0} & 0 & 0.0 & \textcolor{gray}{20.0} & 0 & 0.0 & \textcolor{gray}{20.0} & 0 \\
MP~\citeyearpar{fu2026cap}           & 20.0 & 17.1 & 26 & 0.0 & \textcolor{gray}{30.0} & 0 & 40.0 & 14.6 & 81 & 5.0 & 14.1 & 9 \\
MG~\citeyearpar{mimicgen}            & 0.0 & \textcolor{gray}{30.0} & 0 & \underline{50.0} & 20.9 & 71 & 25.0 & 14.6 & 48 & 20.0 & 12.8 & 39 \\
PGDG~\citeyearpar{dai2026pgdg}       & 0.0 & \textcolor{gray}{30.0} & 0 & \underline{50.0} & 19.1 & \underline{73} & 30.0 & 13.6 & 60 & 20.0 & 14.1 & 38 \\
PLD~\citeyearpar{pld}                & 5.0 & 16.8 & 6 & 30.0 & 17.3 & 41 & 15.0 & 8.3 & 30 & 5.0 & 5.8 & 9 \\
ExpertGen~\citeyearpar{xu2026expertgen} & 30.0 & 17.1 & 41 & 30.0 & 14.5 & 43 & 5.0 & 7.2 & 9 & 5.0 & 5.6 & 9 \\
P-RFCL                               & \underline{45.0} & \underline{9.1} & \underline{79} & 35.0 & \textbf{6.0} & 58 & \underline{45.0} & \textbf{2.8} & \underline{132} & \underline{30.0} & \textbf{3.2} & \underline{72} \\
\midrule
\textbf{Ours}                        & \textbf{90.0} & \textbf{8.7} & \textbf{300} & \textbf{75.0} & \underline{9.3} & \textbf{187} & \textbf{80.0} & \underline{3.4} & \textbf{429} & \textbf{65.0} & \underline{5.1} & \textbf{227} \\
\bottomrule
\end{tabular}}
\end{table}

\begin{table}[t]
\centering
\caption{Simulation experiments.}
\label{tab:main-sim}
\small
\setlength{\tabcolsep}{2.5pt}
\resizebox{\textwidth}{!}{%
\begin{tabular}{@{}lcccccccccccc@{}}
\toprule
& \multicolumn{3}{c}{PnP-Banana} & \multicolumn{3}{c}{StackCube} & \multicolumn{3}{c}{FMB-Round} & \multicolumn{3}{c}{FMB-SqCircle} \\
\cmidrule(lr){2-4}\cmidrule(lr){5-7}\cmidrule(lr){8-10}\cmidrule(lr){11-13}
\textbf{Data source} & SR (\%) $\uparrow$ & Time (s) $\downarrow$ & Succ./h $\uparrow$ & SR (\%) $\uparrow$ & Time (s) $\downarrow$ & Succ./h $\uparrow$ & SR (\%) $\uparrow$ & Time (s) $\downarrow$ & Succ./h $\uparrow$ & SR (\%) $\uparrow$ & Time (s) $\downarrow$ & Succ./h $\uparrow$ \\
\drule
H20                                  & 1.3\,\std{2.3} & 27.5\,\std{4.3} & 2\,\std{3} & 7.3\,\std{5.0} & 21.2\,\std{1.7} & 9\,\std{6} & 0.0\,\std{0.0} & \textcolor{gray}{20.0} & 0\,\std{0} & 0.0\,\std{0.0} & \textcolor{gray}{20.0} & 0\,\std{0} \\
MP~\citeyearpar{fu2026cap}           & 1.3\,\std{2.3} & 29.4\,\std{1.1} & 2\,\std{3} & 3.3\,\std{3.1} & 22.6\,\std{6.4} & 4\,\std{4} & 6.0\,\std{3.5} & 16.3\,\std{1.6} & 11\,\std{6} & 2.7\,\std{3.1} & 16.7\,\std{2.9} & 5\,\std{6} \\
MG~\citeyearpar{mimicgen}            & 41.3\,\std{16.2} & 21.5\,\std{2.5} & 58\,\std{26} & \underline{48.7}\,\std{11.0} & 18.5\,\std{1.1} & \underline{73}\,\std{20} & 31.3\,\std{4.2} & 12.3\,\std{0.1} & 64\,\std{10} & 35.3\,\std{13.0} & 9.6\,\std{0.6} & 81\,\std{37} \\
PGDG~\citeyearpar{dai2026pgdg}       & 27.3\,\std{5.0} & 20.9\,\std{1.0} & 36\,\std{7} & 42.0\,\std{9.2} & 17.2\,\std{0.2} & 62\,\std{16} & 15.3\,\std{4.2} & 14.7\,\std{1.2} & 29\,\std{8} & 30.7\,\std{9.9} & 14.0\,\std{0.5} & 61\,\std{22} \\
PLD~\citeyearpar{pld}                & 20.0\,\std{3.5} & 23.4\,\std{0.5} & 25\,\std{4} & 37.3\,\std{1.2} & 16.1\,\std{0.8} & 54\,\std{3} & 28.0\,\std{3.5} & 6.3\,\std{3.8} & 63\,\std{12} & 8.0\,\std{3.5} & 11.8\,\std{0.9} & 15\,\std{7} \\
ExpertGen~\citeyearpar{xu2026expertgen} & 5.3\,\std{1.2} & 16.7\,\std{3.1} & 7\,\std{1} & 34.7\,\std{12.2} & 12.7\,\std{0.9} & 53\,\std{23} & 34.0\,\std{9.2} & 10.5\,\std{1.9} & 75\,\std{26} & \underline{36.0}\,\std{8.7} & 8.4\,\std{0.1} & \underline{83}\,\std{25} \\
P-RFCL                               & \underline{73.3}\,\std{12.2} & \textbf{8.4}\,\std{0.9} & \underline{192}\,\std{60} & 28.7\,\std{3.1} & \textbf{5.1}\,\std{0.2} & 45\,\std{6} & \underline{38.7}\,\std{1.2} & \textbf{3.4}\,\std{0.0} & \underline{103}\,\std{5} & 20.7\,\std{5.8} & \underline{4.2}\,\std{0.3} & 45\,\std{14} \\
\midrule
\textbf{Ours}                        & \textbf{85.3}\,\std{6.4} & \underline{9.3}\,\std{0.7} & \textbf{253}\,\std{44} & \textbf{56.7}\,\std{13.0} & \underline{9.3}\,\std{0.2} & \textbf{116}\,\std{45} & \textbf{61.3}\,\std{3.1} & \underline{4.2}\,\std{0.1} & \textbf{215}\,\std{20} & \textbf{44.7}\,\std{4.2} & \textbf{3.6}\,\std{0.1} & \textbf{127}\,\std{18} \\
\bottomrule
\end{tabular}}
\end{table}

\subsection{Diversity trade-off on every task}
\label{sec:alpha-full}

\begin{figure}[t]
\centering
\includegraphics[width=\linewidth]{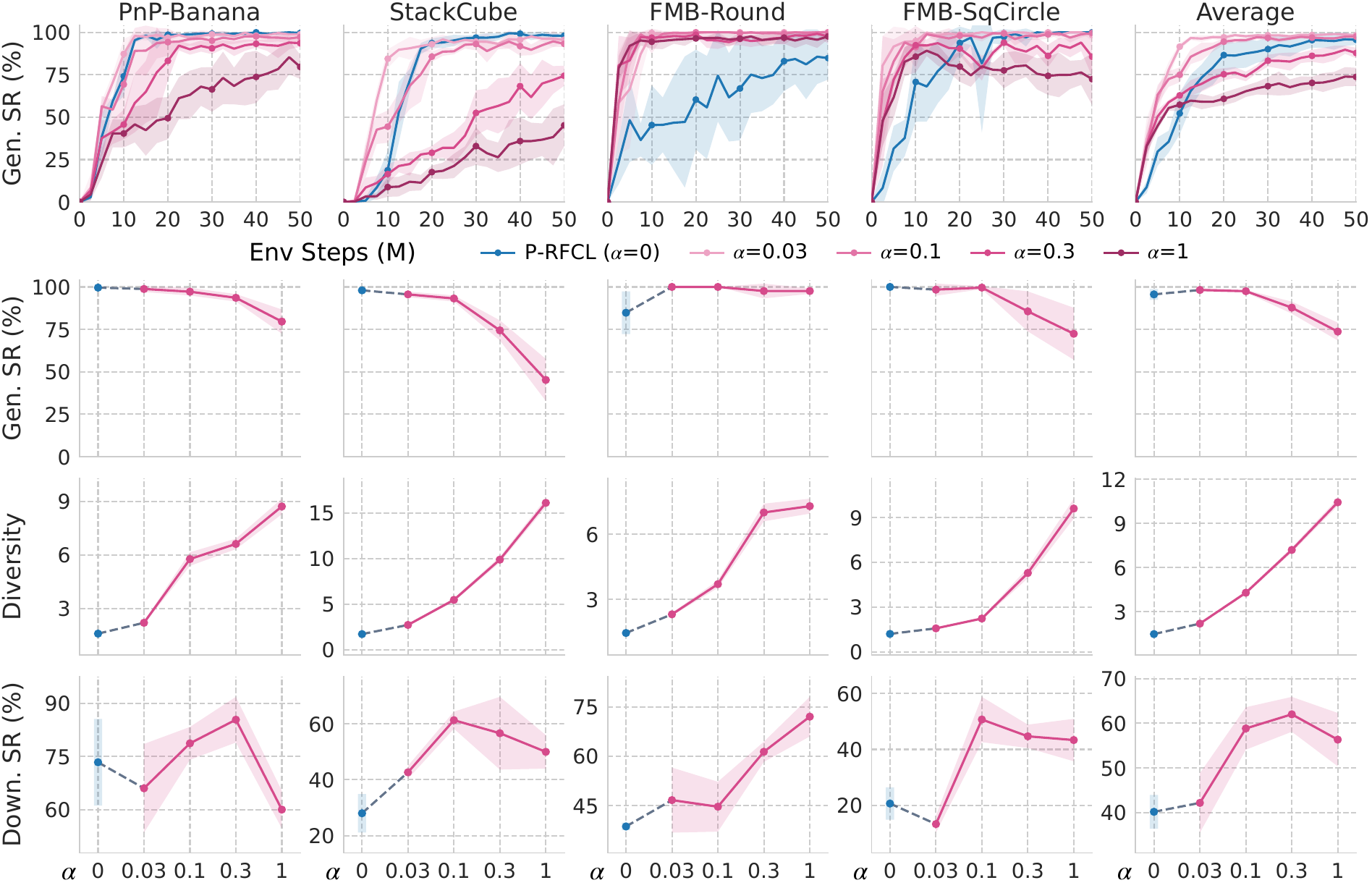}
\caption{%
The shape of the trade-off is common to the four tasks, but the $\alpha$ at which generator success
starts to give way is not.
}
\label{fig:alpha-full}
\end{figure}

\Cref{fig:alpha-tradeoff} combines PnP-Banana trajectories with four-task averages; \Cref{fig:alpha-full} reports generator success, diversity, and downstream success per task and on average.
The top row adds the generator's training curve at each $\alpha$, and the generator success below
is where each curve ends; both average five seeds per cell and shade $\pm$ one standard deviation.

\subsection{Qualitative trajectories on StackCube and FMB-Round}
\label{sec:appn-qualitative}

\Cref{fig:qualitative} shows PnP-Banana and FMB-SqCircle; the remaining two tasks, drawn the same way, are below.

\subsection{Efficiency Measurement}
\label{sec:appn-efficiency}
\paragraph{Generator training.}
The wall-clock training time of P-RFCL and \method{} in \Cref{fig:rl-baselines} is estimated from measurements on one B200 GPU with in-training evaluation, checkpointing, and video recording disabled: we time $1.5$M environment steps of each curriculum phase and extrapolate to $50$M steps using each run's own phase-switch step, so that queueing and preemption do not enter the comparison.
PLD and ExpertGen report the wall-clock time of their own training runs on the same GPU type.
Every curve in \Cref{fig:rl-baselines} averages five training seeds, with the shaded band showing $\pm1$ standard deviation.

\paragraph{End-to-end cost.}
The GPU-hours in \Cref{fig:dataset-analysis}~(d) are normalized to a single process on one GPU, so that job packing and preemption do not enter.
Generator training counts the full $50$M-step run without in-training evaluation, and data generation counts the rollout and rendering of $3{,}000$ retained demonstrations; $\pi_{0.5}$ fine-tuning is excluded because it is identical for every source.
The per-task split into the two stages is the bottom row of \Cref{fig:dataset-taskwise}.

\subsection{Qualitative Rollouts}
\label{sec:appn-rollouts}
We show how each method solves every task over time, first for the data generators and then for the downstream policies trained on their data.
Each figure has one row per data source and one column per time step, from $0$ to $28$\,s for PnP-Banana and StackCube and to $17.5$\,s for the insertion tasks; the interval is shared by all rows and by both kinds of figures, so a row ends earlier when its method completes the task sooner.
When a rollout completes, the next cell shows its final frame together with the time at which it completed; a rollout that does not complete within the shown horizon is marked as a failure in the last column.
Full videos are available on the project page.

\begin{figure}[htbp]
\centering
\includegraphics[width=\linewidth]{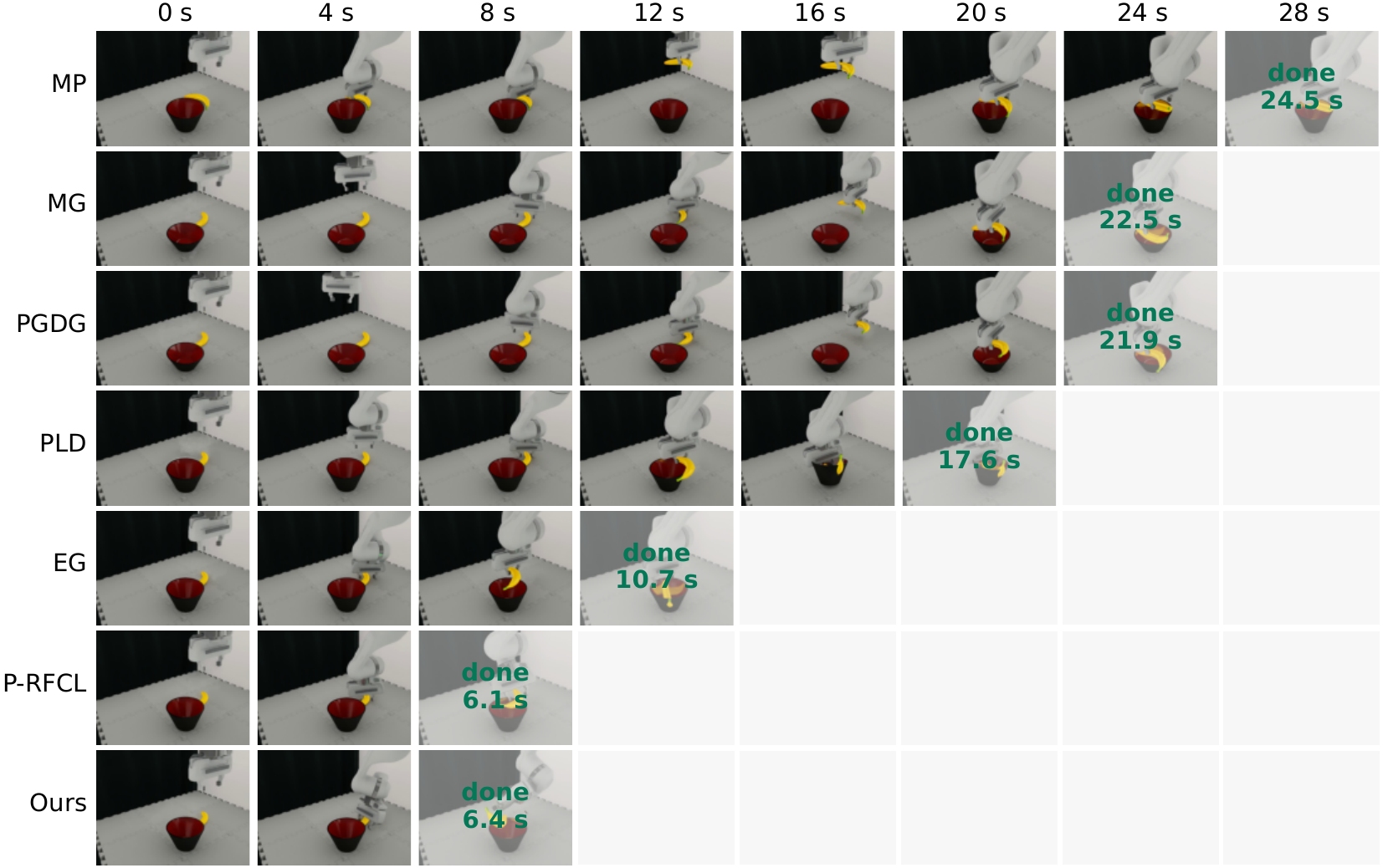}
\vspace{-0.4cm}
\caption{\textbf{Data generators on PnP-Banana.} Rollouts from the shared initial scene.}
\label{fig:quali-teacher-banana}
\end{figure}

\begin{figure}[htbp]
\centering
\includegraphics[width=\linewidth]{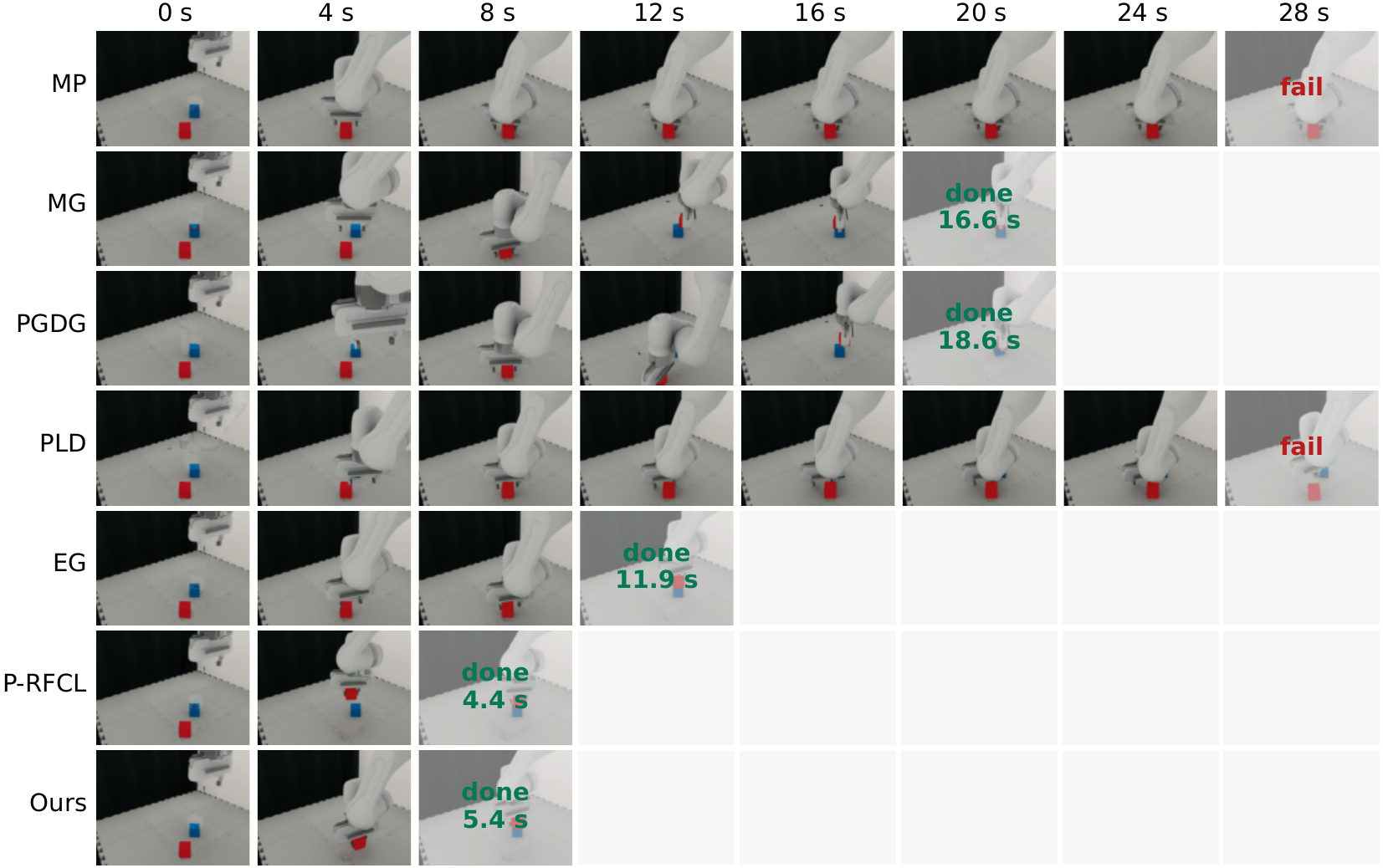}
\vspace{-0.4cm}
\caption{\textbf{Data generators on StackCube.} Rollouts from the shared initial scene.}
\label{fig:quali-teacher-cube}
\end{figure}

\begin{figure}[htbp]
\centering
\includegraphics[width=\linewidth]{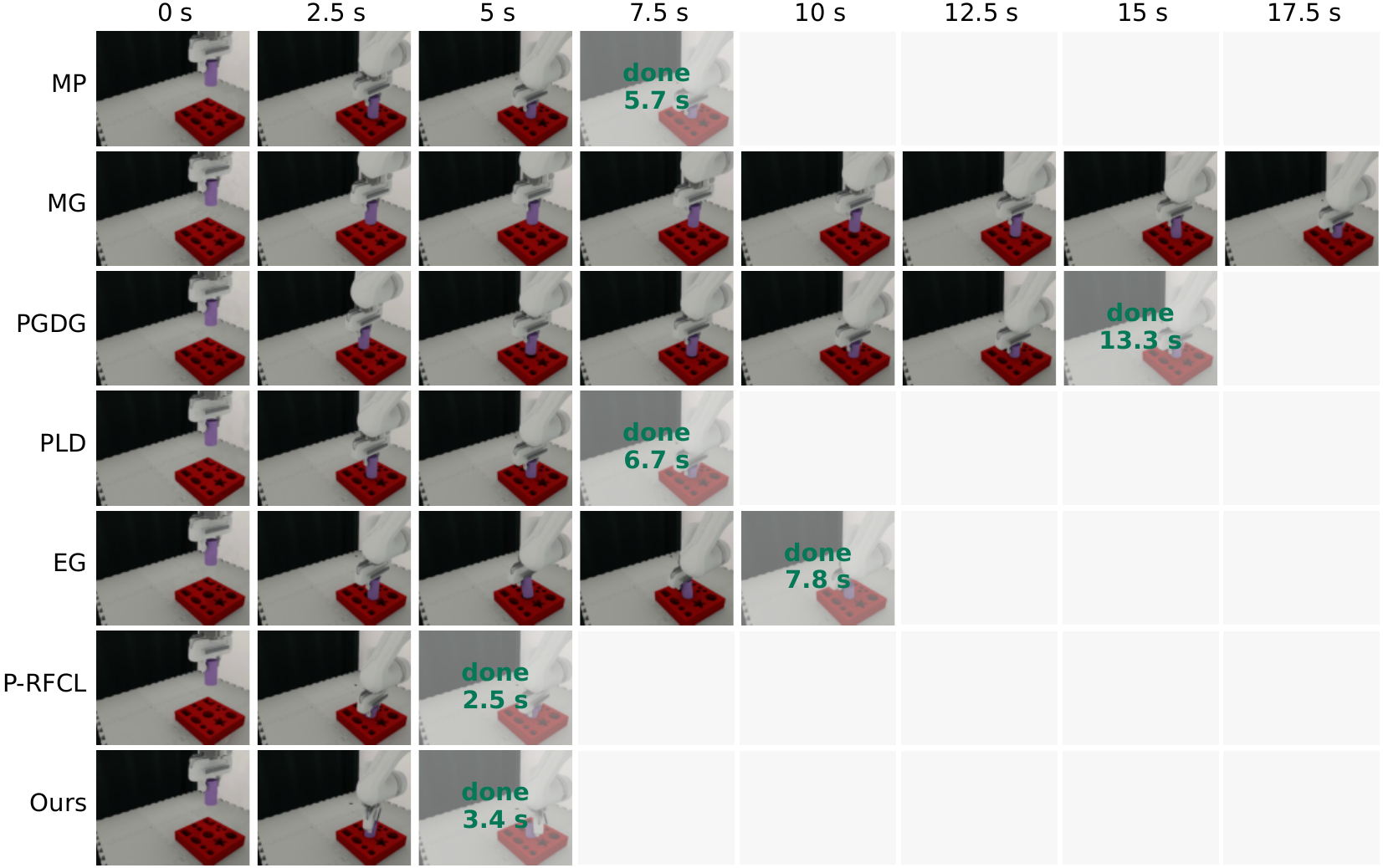}
\vspace{-0.4cm}
\caption{\textbf{Data generators on FMB-Round.} Rollouts from the shared initial scene.}
\label{fig:quali-teacher-round}
\end{figure}

\begin{figure}[htbp]
\centering
\includegraphics[width=\linewidth]{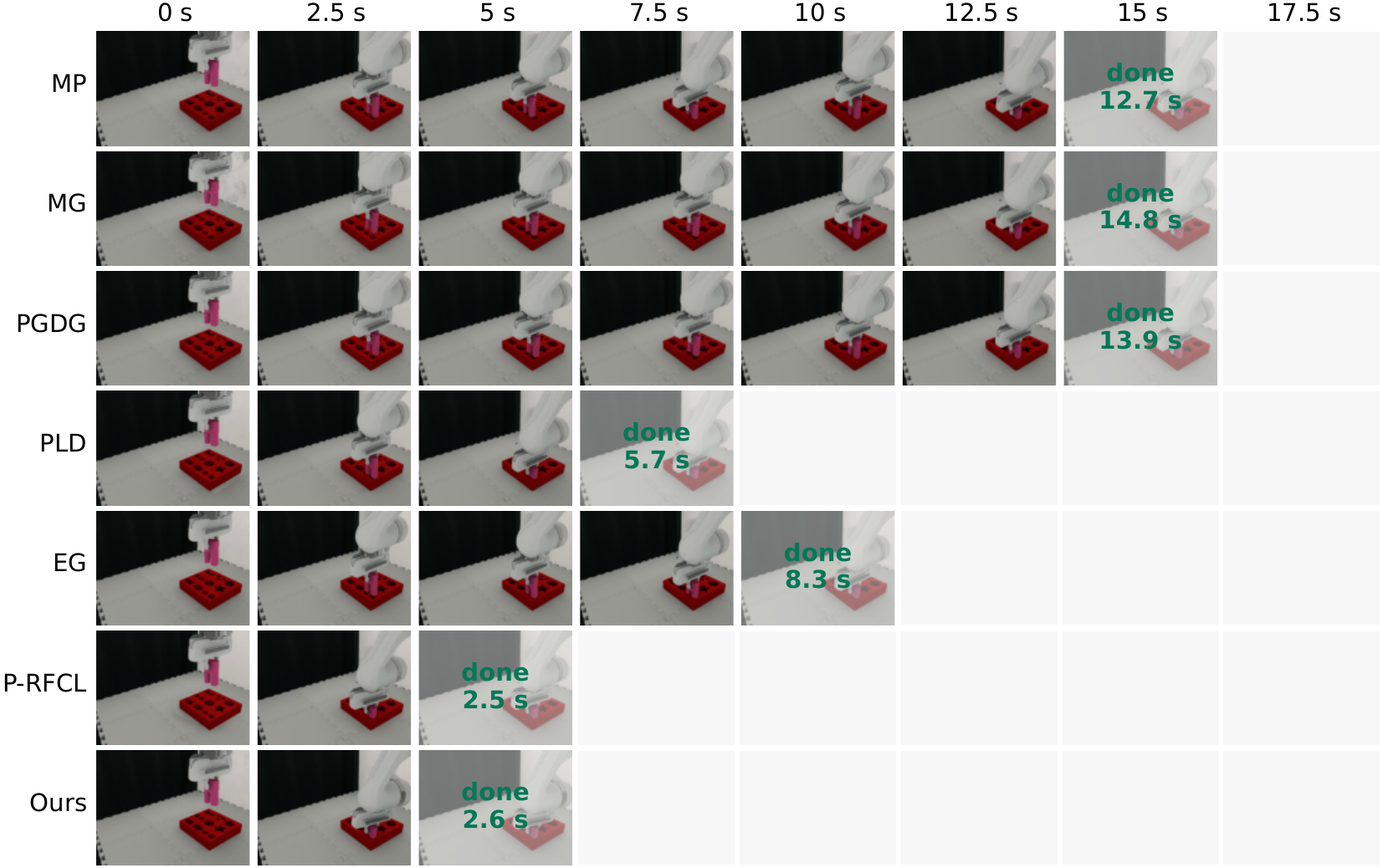}
\vspace{-0.4cm}
\caption{\textbf{Data generators on FMB-SqCircle.} Rollouts from the shared initial scene.}
\label{fig:quali-teacher-sqcirc}
\end{figure}

\paragraph{Data generators.}
\Cref{fig:quali-teacher-banana,fig:quali-teacher-cube,fig:quali-teacher-round,fig:quali-teacher-sqcirc} roll out each generator from the same fixed initial scene per task as \Cref{fig:qualitative}.
Each row shows the successful rollout whose length is closest to the average episode length of that generator's dataset (\Cref{fig:dataset-analysis}); on StackCube, MP and PLD produced no successful rollout from this scene, so a failed one is shown instead.
H20 has no generator and is therefore omitted.

\begin{figure}[htbp]
\centering
\includegraphics[width=\linewidth]{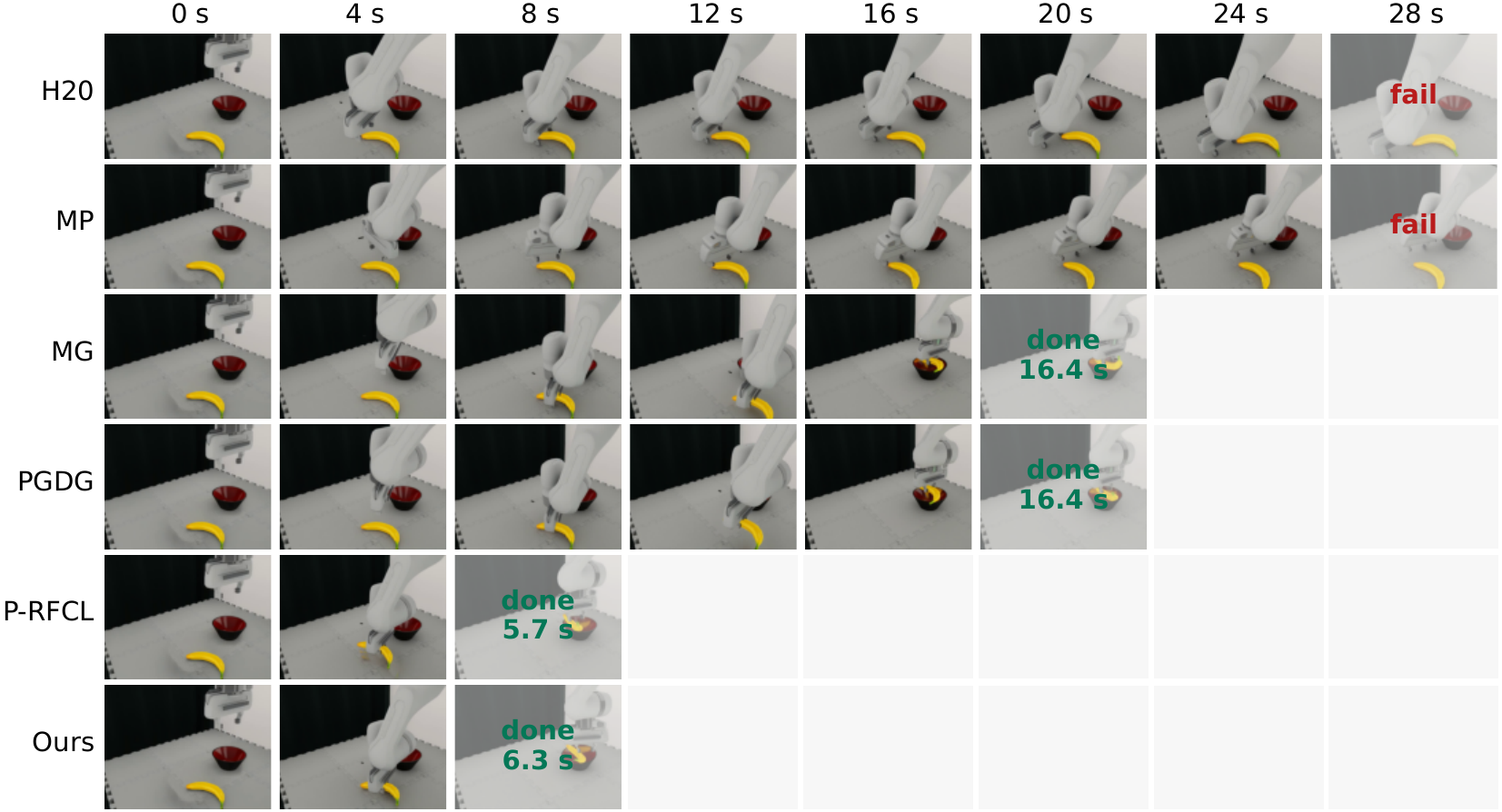}
\vspace{-0.4cm}
\caption{\textbf{Downstream policies on PnP-Banana.} Rollouts on the selected shared initial state.}
\label{fig:quali-sft-banana}
\end{figure}

\begin{figure}[htbp]
\centering
\includegraphics[width=\linewidth]{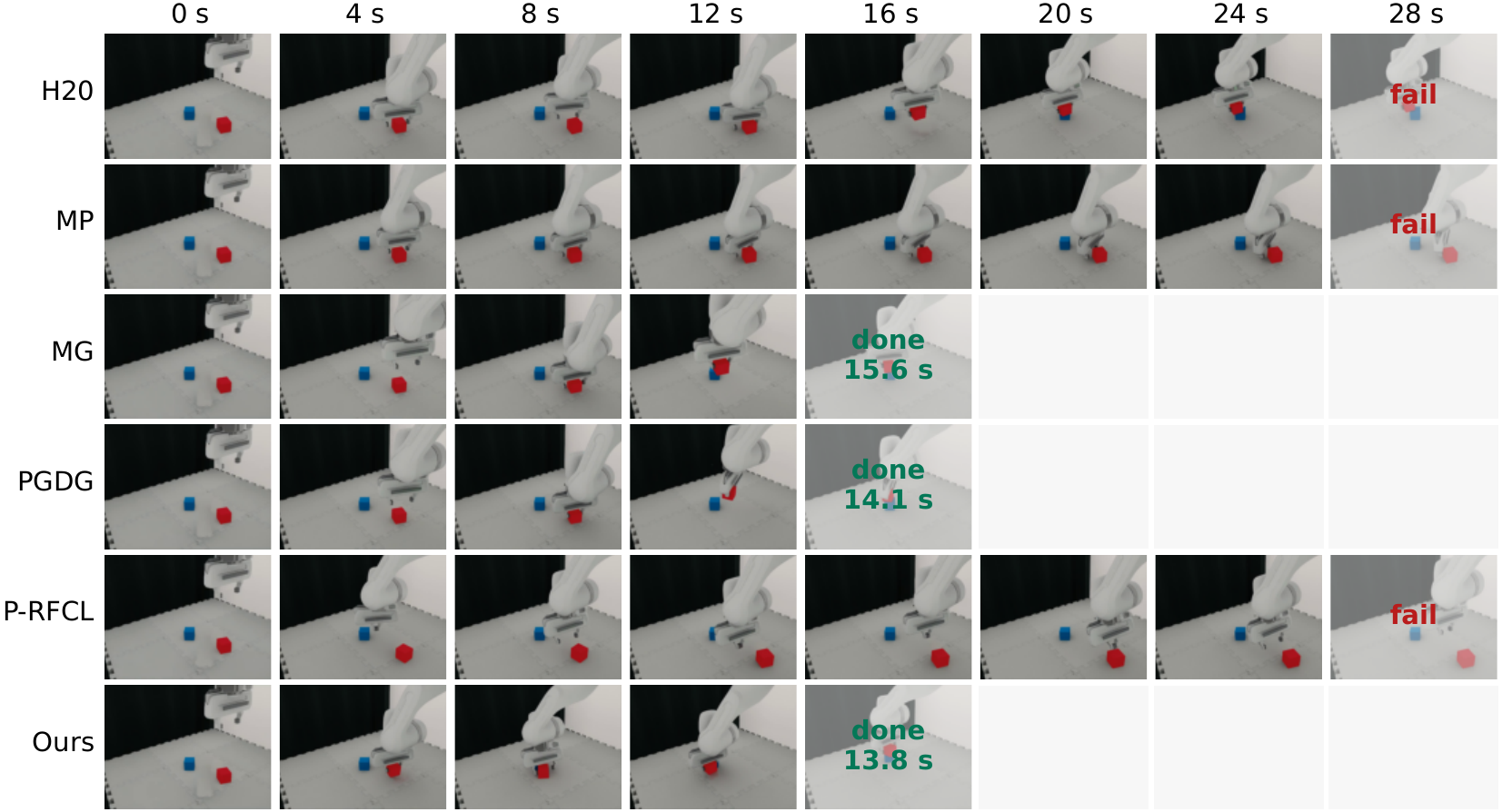}
\vspace{-0.4cm}
\caption{\textbf{Downstream policies on StackCube.} Rollouts on the selected shared initial state.}
\label{fig:quali-sft-cube}
\end{figure}

\begin{figure}[htbp]
\centering
\includegraphics[width=\linewidth]{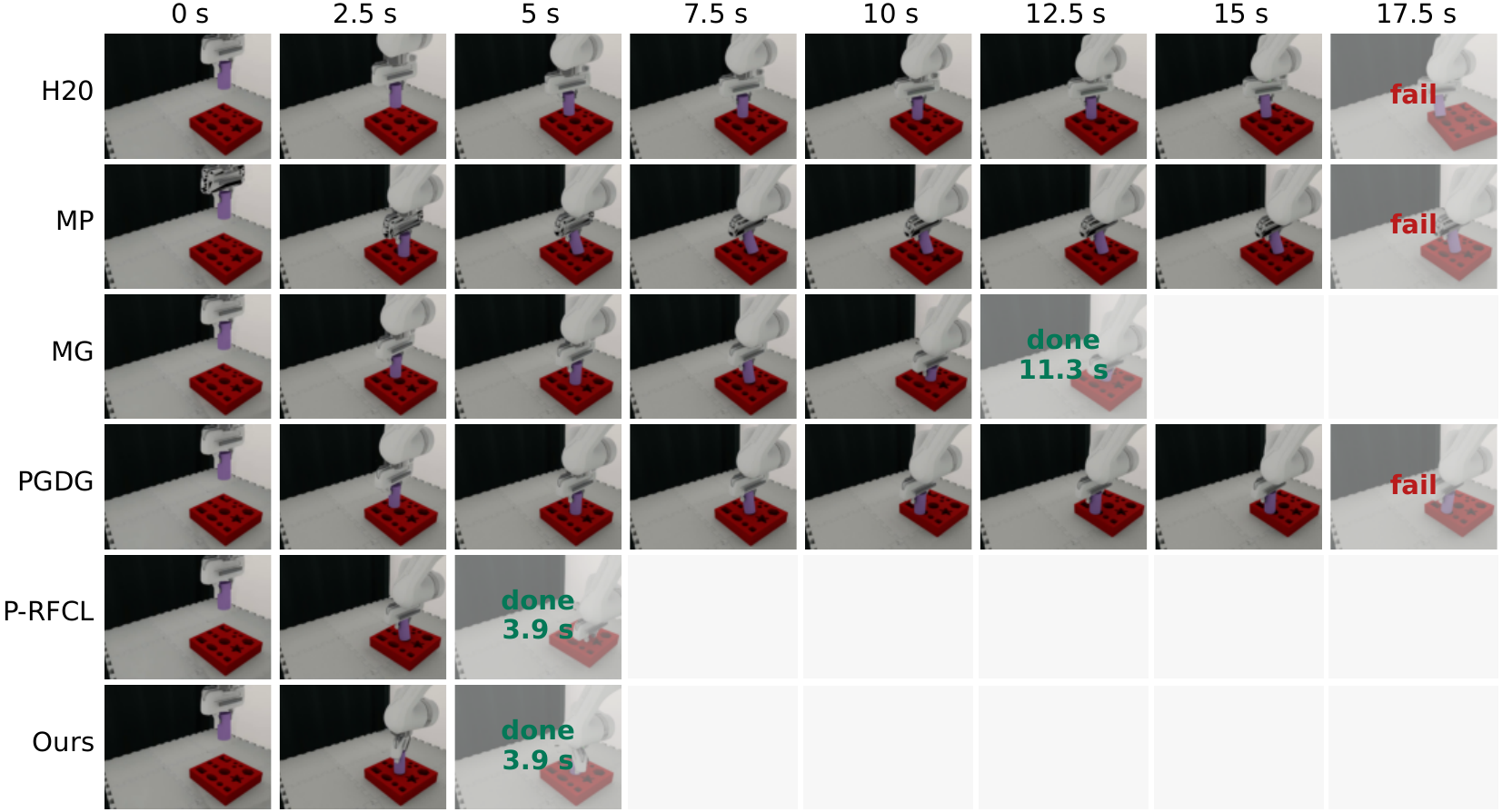}
\vspace{-0.4cm}
\caption{\textbf{Downstream policies on FMB-Round.} Rollouts on the selected shared initial state.}
\label{fig:quali-sft-round}
\end{figure}

\begin{figure}[htbp]
\centering
\includegraphics[width=\linewidth]{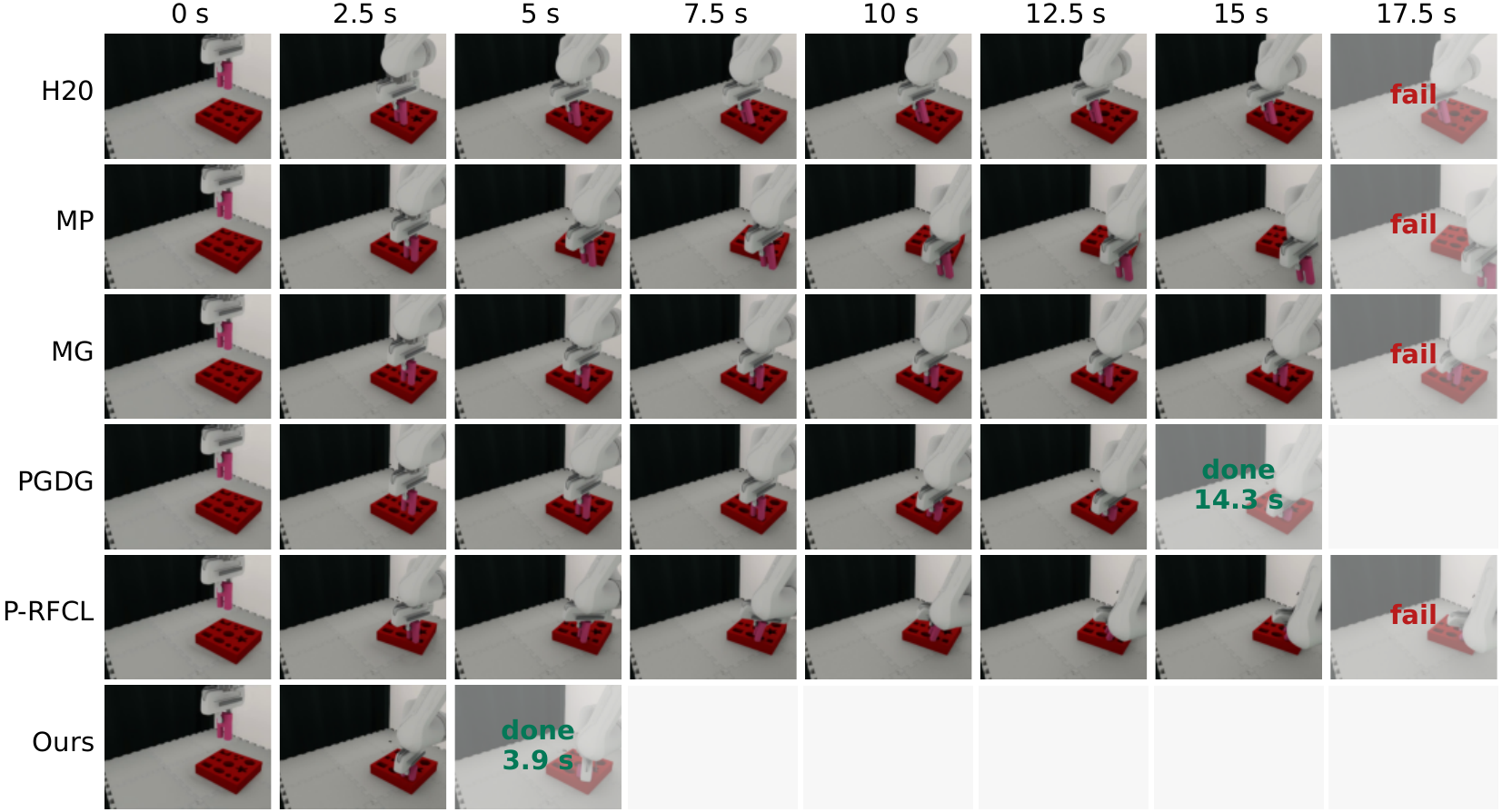}
\vspace{-0.4cm}
\caption{\textbf{Downstream policies on FMB-SqCircle.} Rollouts on the selected shared initial state.}
\label{fig:quali-sft-sqcirc}
\end{figure}

\paragraph{Downstream policies.}
\Cref{fig:quali-sft-banana,fig:quali-sft-cube,fig:quali-sft-round,fig:quali-sft-sqcirc} show the $\pi_{0.5}$ policies fine-tuned on each dataset, all rolled out in simulation from the same initial state.
We evaluate every policy on ten initial states shared by all methods and show, for each task, the one on which the most methods succeed.
A row ends when the policy first satisfies the success condition.
PLD and ExpertGen are omitted from these figures.

\begin{figure}[htbp]
\centering
\includegraphics[width=\linewidth]{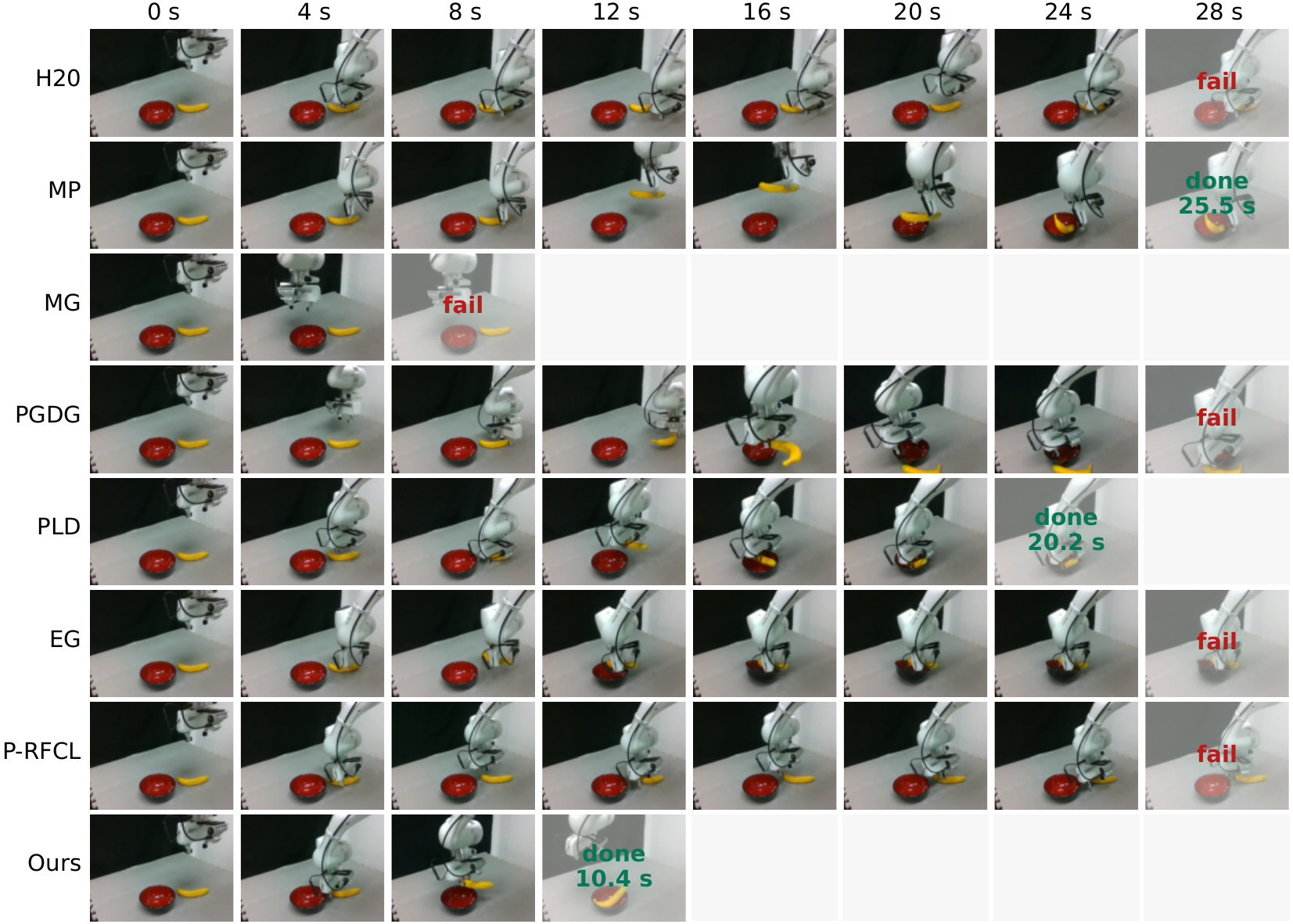}
\vspace{-0.4cm}
\caption{\textbf{Real-robot rollouts on PnP-Banana.} Rollouts on the scene used in the project-page video.}
\label{fig:quali-real-banana}
\end{figure}

\begin{figure}[htbp]
\centering
\includegraphics[width=\linewidth]{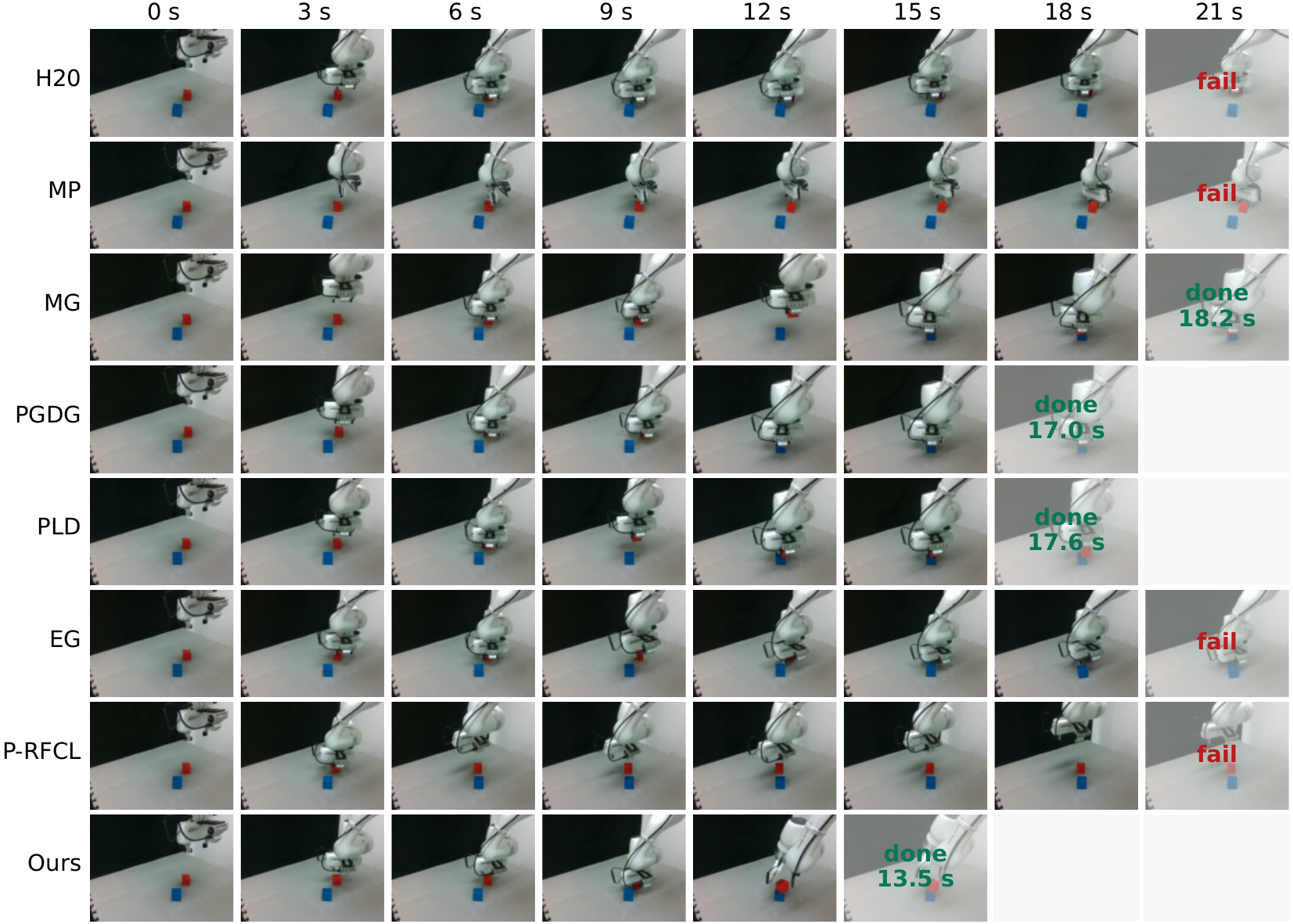}
\vspace{-0.4cm}
\caption{\textbf{Real-robot rollouts on StackCube.} Rollouts on the scene used in the project-page video.}
\label{fig:quali-real-cube}
\end{figure}

\begin{figure}[htbp]
\centering
\includegraphics[width=\linewidth]{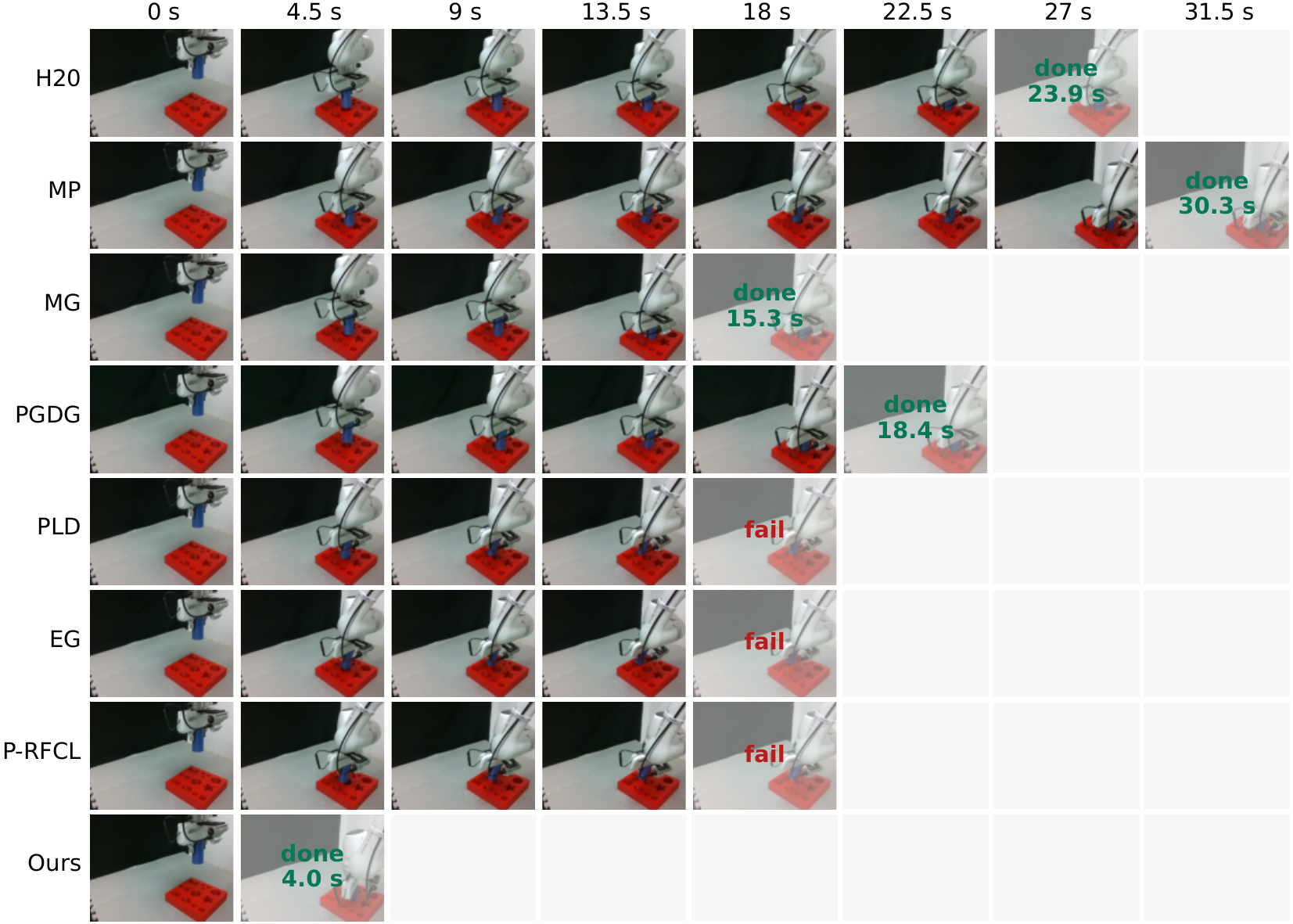}
\vspace{-0.4cm}
\caption{\textbf{Real-robot rollouts on FMB-Round.} Rollouts on the scene used in the project-page video.}
\label{fig:quali-real-round}
\end{figure}

\begin{figure}[htbp]
\centering
\includegraphics[width=\linewidth]{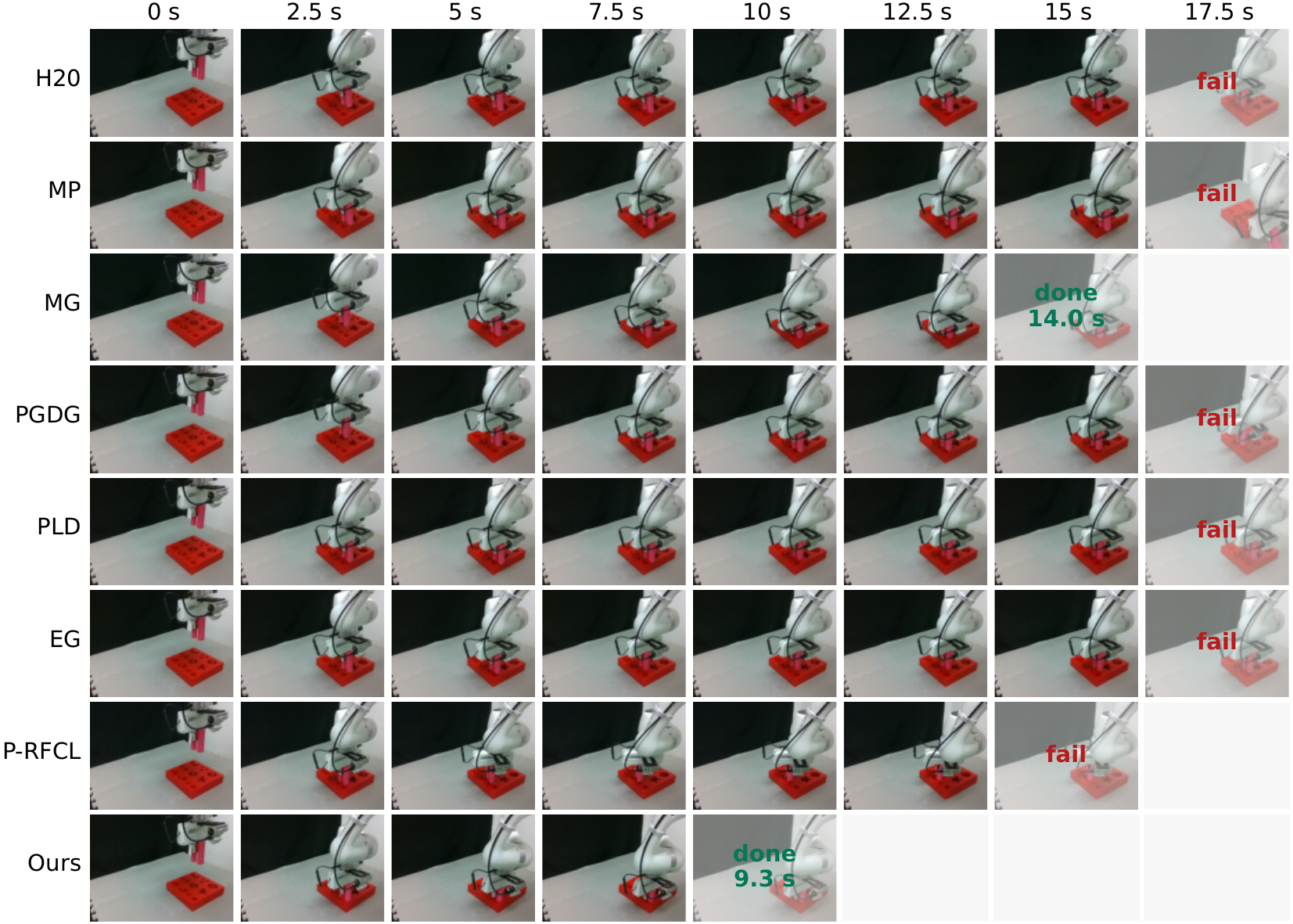}
\vspace{-0.4cm}
\caption{\textbf{Real-robot rollouts on FMB-SqCircle.} Rollouts on the scene used in the project-page video.}
\label{fig:quali-real-sqcirc}
\end{figure}

\paragraph{Real robot.}
\Cref{fig:quali-real-banana,fig:quali-real-cube,fig:quali-real-round,fig:quali-real-sqcirc} show the same downstream policies on the real robot, including PLD and ExpertGen.
For each task we reproduce three initial scenes for every method, and show the one used to compare completion speed in the project-page video; time starts at the first policy command.
A successful run ends at the time it completed, and a run that ends without success is marked as a failure in the next column.

\end{document}

%% file: math_commands.tex
\usepackage{amsmath,amsfonts,bm}

\def\eqref#1{equation~\ref{#1}}

\def\1{\bm{1}}

\DeclareMathAlphabet{\mathsfit}{\encodingdefault}{\sfdefault}{m}{sl}
\SetMathAlphabet{\mathsfit}{bold}{\encodingdefault}{\sfdefault}{bx}{n}

\def\gA{{\mathcal{A}}}

\def\gS{{\mathcal{S}}}
\def\gT{{\mathcal{T}}}

\newcommand{\E}{\mathbb{E}}

\DeclareMathOperator*{\argmin}{arg\,min}

%% file: preamble.tex
\usepackage{graphicx}   
\usepackage[dvipsnames]{xcolor}

\definecolor{rose}{RGB}{214, 74, 139}
\newcommand{\ourc}[1]{\textcolor{rose}{#1}}
\definecolor{hgreen}{RGB}{28, 164, 98}

\usepackage{kotex}
\usepackage{lipsum}
\usepackage{pifont}
\usepackage[normalem]{ulem}
\newcommand{\cmark}{\ding{51}}%
\newcommand{\xmark}{\ding{55}}%
\usepackage{mathtools}

\usepackage{array}
\newcolumntype{P}[1]{>{\centering\arraybackslash}p{#1}}
\newcolumntype{M}[1]{>{\centering\arraybackslash}m{#1}}
\newcolumntype{L}[1]{>{\hspace{0.5em}\raggedright\arraybackslash}m{#1}}
\newcolumntype{R}[1]{>{\raggedleft\arraybackslash}m{#1}}
\newcolumntype{W}[1]{>{\raggedright\arraybackslash}m{#1}}  

\usepackage{xspace}

\makeatletter
\DeclareRobustCommand\onedot{\futurelet\@let@token\@onedot}
\def\@onedot{\ifx\@let@token.\else.\null\fi\xspace}

\makeatother

\usepackage[colorlinks=true, linkcolor=blue, citecolor=blue, urlcolor=blue, pagebackref=true]{hyperref}
\definecolor[named]{ACMDarkBlue}{cmyk}{1,0.58,0,0.21}
\definecolor{darkgreen}{RGB}{25,200,25}
\colorlet{citerose}{rose!75!white}  
\hypersetup{colorlinks,
  linkcolor=citerose,  
  citecolor=citerose,  
  urlcolor=citerose,   
  filecolor=citerose
}
\renewcommand*{\backref}[1]{}  
\renewcommand*{\backrefalt}[4]{%
  \ifcase #1 %
    (Not cited.)%
  \or
    (Cited on page #2.)%
  \else
    (Cited on pages #2.)%
  \fi
}

\usepackage{cleveref}
\Crefformat{figure}{Fig.~#2#1#3}
\Crefformat{equation}{(#2#1#3)}
\Crefformat{table}{Tab.~#2#1#3}

\AtBeginDocument{\let\cite\citep}

\usepackage{caption}
\DeclareCaptionFormat{tablewithgap}{#1#2\vspace{3pt}#3}

\usepackage{tabularx,booktabs}
\newcolumntype{Y}{>{\centering\arraybackslash}X}
\usepackage{colortbl}
\usepackage{multirow}
\usepackage{soul}
\usepackage{wrapfig}
\newcommand{\drule}{\specialrule{0.2pt}{1pt}{1pt}%
\specialrule{0.2pt}{0pt}{\belowrulesep}%
}

\newcommand{\std}[1]{{\scriptsize$\pm$#1}}

\usepackage{footmisc}
\newcommand{\dagfootnote}[1]{%
  \textsuperscript{\dag}
  \begingroup
    \let\savedfootnoterule\footnoterule
    \renewcommand{\thefootnote}{}
    \renewcommand{\footnoterule}{
      \vspace{1ex}\hrule width 0.4\textwidth\vspace{1ex}%
    }%
    \footnotetext{\dag\ #1}
    \addtocounter{footnote}{-1}
    \let\footnoterule\savedfootnoterule
  \endgroup
}

\usepackage{algorithm,algpseudocode}
\usepackage[algo2e]{algorithm2e} 
\definecolor{commentcolor}{RGB}{110,154,155}
\definecolor{defcolor}{RGB}{225,81,145}

\usepackage{titletoc}



\usepackage{enumitem}
\usepackage{flushend}

\definecolor{reviewblue}{RGB}{45,85,155}

\definecolor{analysisteal}{RGB}{0,125,115}